\documentclass{article}
\usepackage{ijcai26}

\usepackage{times}
\usepackage{soul}
\usepackage{url}
\usepackage[hidelinks]{hyperref}
\usepackage[utf8]{inputenc}
\usepackage[small]{caption}
\usepackage{graphicx}
\usepackage{amsmath, amssymb}
\usepackage{amsthm}
\usepackage{booktabs}
\usepackage{algorithm}
\usepackage{algorithmic}
\usepackage[switch]{lineno}

\usepackage{booktabs}    
\usepackage{multirow}    
\usepackage{makecell}    
\usepackage{colortbl}    
\usepackage{xcolor}      
\usepackage{adjustbox}   
\usepackage{dsfont}

\usepackage{pifont}
\newcommand{\cmark}{\ding{51}} 

\title{SpatialSV: Internalizing Interpretable 3D Spatial Awareness in MLLMs via Task-Oriented Visual Supervision}

\author{
    Jiayu Tang\and
    Yuchen Zhou\and
    Chao Gou\thanks{Corresponding authors}
    \affiliations
    School of Intelligent Systems Engineering, Sun Yat-sen University
    \emails
    tangjy59@mail2.sysu.edu.cn, zhouych37@mail2.sysu.edu.cn, gouchao@mail.sysu.edu.cn
}

\begin{document}

\maketitle

\begin{abstract}
    Unlocking the spatial intelligence of multimodal large language models (MLLMs) is crucial for understanding and interacting with the 3D world. 
    Prevailing approaches typically inject spatial priors via external tools, which impose significant inference overhead, or rely on latent feature distillation, which remains uninterpretable and lacks fine-grained geometric constraints.
    To address these issues, we propose SpatialSV, a framework designed to internalize robust 3D spatial awareness within MLLMs while simultaneously offering inherent interpretability. 
    Deviating from passive feature imitation, SpatialSV employs task-oriented visual supervision, compelling the model to actively lift its 2D visual features into explicit 3D representations, including depth maps, camera poses, and point clouds. 
    Crucially, this 2D-to-3D lifting process provides a transparent window into the model’s representations: the resulting 3D reconstructions serve as an intuitive proxy for visualizing and diagnosing the quality of the model’s intrinsic spatial knowledge.
    Extensive experiments across multiple models and benchmarks 
    demonstrate the effectiveness of SpatialSV in enhancing and interpreting MLLMs’ spatial intelligence. 
    Furthermore, the framework exhibits strong generalization in semi-supervised settings, validating its potential to leverage unlabeled visual data for scalable, interpretable spatial representation learning.
\end{abstract}

\section{Introduction}
Spatial intelligence refers to the ability to understand, interpret and reason about the spatial relationships and properties of objects and scenes. This ability is fundamental for real-world tasks ranging from autonomous driving \cite{zhou2025towards,tang2026letp} to robotic manipulation \cite{zhou2023learning,song2025robospatial}, where understanding and interacting with the 3D environments is essential. Despite significant progress in spatial reasoning and scene understanding, even the most advanced multimodal large language models (MLLMs) face challenges in maintaining cross-view consistency and reasoning about occluded objects \cite{yin2025spatial,yang2025thinking}. These limitations highlight a fundamental weakness of MLLMs—the reliance on 2D visual-textual data and autoregressive training paradigms—which fails to yield robust and consistent internal 3D spatial representations. This absence of 
intrinsic 3D spatial awareness
is a major bottleneck toward genuine spatial intelligence. 

\begin{figure}[!t]
\captionsetup{skip=0pt}
\centering
    \includegraphics[width=0.485\textwidth]{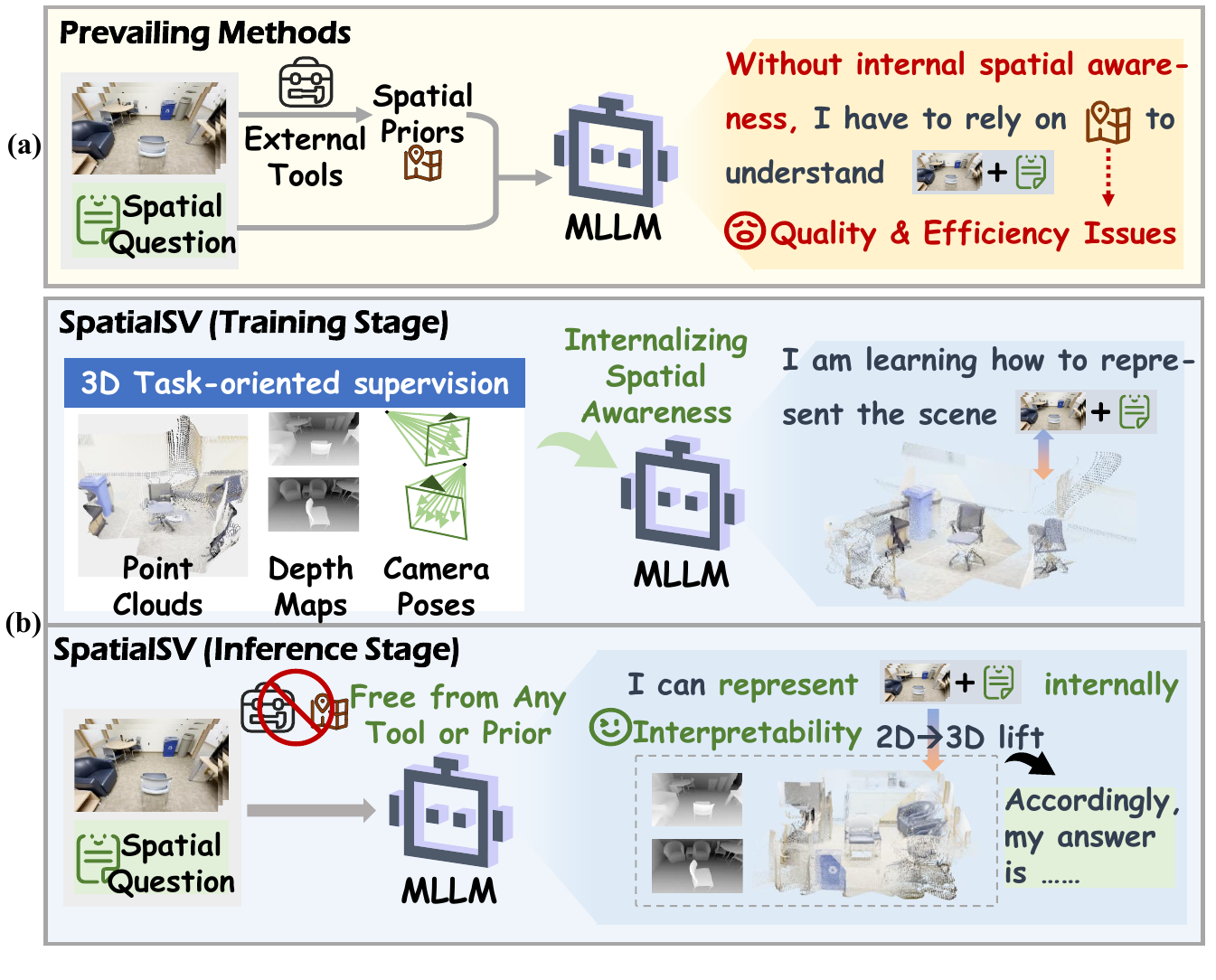}
    \caption{(a) Prevailing methods incorporate external spatial priors as input to MLLMs, suffering from prior quality and inference efficiency issues. 
    (b) SpatialSV internalizes 3D spatial awareness in MLLMs via task-oriented visual supervision, which enables spatial understanding with interpretable intrinsic spatial representations.}
    \label{Introduction_fig}
\end{figure}

To address this limitation, 
one major line of research focuses on providing MLLMs with external spatial priors to compensate for their insufficient intrinsic representations, as illustrated in Figure \ref{Introduction_fig} (a).
Among these, prompt-based methods rely on external models or tools to construct prompts enriched with spatial priors, 
\cite{yang2025mindjourney,li2025see,qi2025gpt4scene,zhu2025struct2d,liu2025abstract}. However, such methods introduce substantial inference overhead and are sensitive to errors from external models. 
A more direct alternative is to explicitly encode 3D data such as point clouds and depth maps, and project them to the language-aligned space of large language models \cite{chen2024ll3da,li2025spatial,wang2025spatial,zhu2025llava}. However, these methods typically depend on highly customized encoding and alignment modules and will introduce both training and inference complexity. Moreover, the strong 2D pretraining bias of MLLMs poses an additional obstacle to the explicit integration of 3D data.

Instead of emphasizing the construction and utilization of external spatial representations, we argue that MLLMs should internalize spatial awareness to form an internal spatial mental model \cite{johnson1980mental,yin2025spatial}. 
Imaging in a practical embodied AI system where acquiring 3D data online is often costly and challenging with strict real‑time constraints, MLLMs must rely on their own spatial awareness to construct internal representations of the environment, while maintaining a lightweight architecture to enable fast inference. 
To this end, recent works attempt to inject spatial awareness into MLLMs by distilling features from 3D vision foundation models (VFMs) \cite{huang2025mllms,chen2025think}. 
However, \textit{the representation learned via distillation is inherently hard to interpret}, limiting a deeper understanding of 
the internal spatial mental modeling mechanism \cite{johnson1983mental} within MLLMs.

\begin{figure}[t]
\centering
    \includegraphics[width=0.45\textwidth]{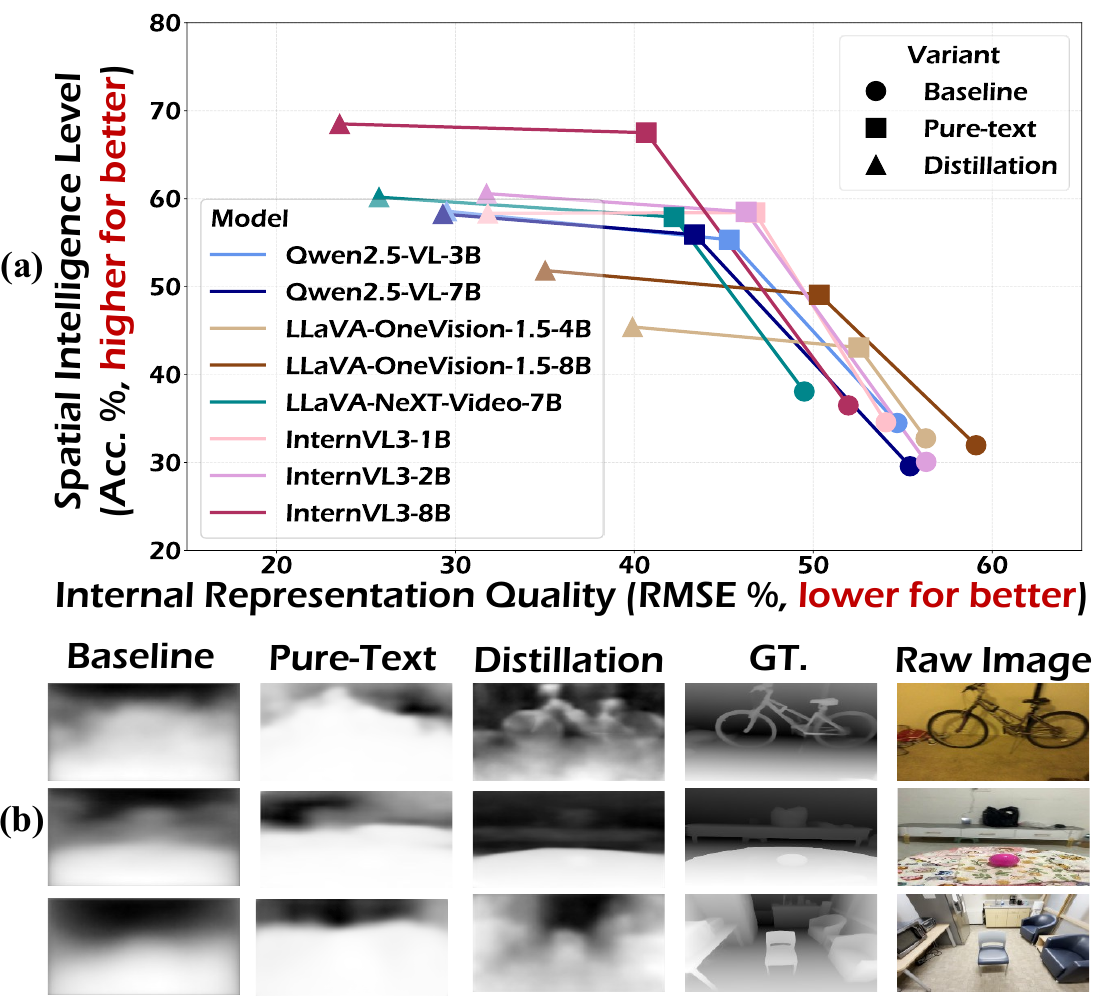}
    \caption{Depth probing results. (a) Quantitative results: the correlation between the quality of internal representations and the level of spatial intelligence. We compare 8 MLLMs under three variants: an untuned baseline, a text supervision variant, and a distillation variant. (b)  Qualitative results: visualized depth maps of three variants for Qwen2.5-VL-3B, along with the ground-truths and raw images.}
    \label{Depth_Prob}
\end{figure}

Inspired by the probing techniques \cite{el2024probing,chen2025feat2gs}, we conduct a pilot study to investigate the quality of spatial representations across different MLLMs under two paradigms: pure autoregressive text supervision and feature distillation. 
The probing results, as illustrated in Figure \ref{Depth_Prob}, reveal the following observations: 
(1) The quality of an MLLM’s intrinsic spatial representation is positively correlated with its level of spatial intelligence. 
(2) Introducing 3D-aware supervision improves the quality of spatial representations. 
These findings validate the effectiveness of 3D‑aware supervision in enhancing the spatial intelligence of MLLMs. 
However, the performance gap between the depth probing results and the ground‑truth further exposes the limitations of the distillation paradigm. This is because \textit{distilling features from 3D VFMs constitutes a coarse‑grained alignment process that lacks explicit and structured spatial constraints}. In addition, aligning feature dimensions during distillation inevitably incurs information loss in the target representations, further weakening the robustness of the supervision signal. 

These insights raise a central question: \textit{can we identify a more intuitive and fine-grained form of 3D-aware supervision?} In response to this question, we propose SpatialSV, a framework that internalizes interpretable and robust 3D spatial awareness in MLLMs via task-oriented visual supervision, as illustrated in Figure \ref{Introduction_fig} (b). 
Specifically, we perform 2D-to-3D lifting of MLLMs' visual features and align them with explicit spatial representations such as depth maps and point clouds, so as to unlock their intrinsic spatial intelligence.
Inspired by 3D VFMs which leverage overlapping yet complementary downstream 3D tasks to learn geometrically consistent representations, we incorporate a set of complementary 3D tasks, including depth estimation, point‑cloud reconstruction, and ray‑map prediction. 
Technically, we extract multi‑layer hidden visual features from the MLLM and employ projection layers together with decoupled DPT modules to perform multi‑task prediction. 
By combining the coarse‑grained guidance of feature distillation with fine‑grained, task‑oriented constraints, SpatialSV facilitates robust intrinsic spatial representation learning of MLLMs. 
Extensive experiments across multiple MLLMs and benchmarks demonstrate the superior performance of SpatialSV in enhancing spatial intelligence.

Additionally, the 3D lifting results derived from MLLMs serve as an intuitive and interpretable manifestation of models' internal spatial representations. 
Through both quantitative and qualitative analyses, we investigate a distinct correlation between 3D lifting results and intra-model spatial intelligence as well as sample difficulties, highlighting the inherent interpretability of SpatialSV. Furthermore, we apply SpatialSV to a semi‑supervised setting where $50\%$ of the samples have no textual annotations and achieve performance comparable to full text supervision. This is particularly important for scenarios where 3D text annotations are scarce.
In summary, our contributions include:
\begin{itemize}
     \item We propose SpatialSV, a novel framework that internalizes robust 3D spatial awareness in MLLMs through task-oriented visual supervision, with the 3D lifting results serving as an intuitive and interpretable proxy for intra-model spatial representations.

    \item We introduce a probe-based representation analysis framework for MLLMs, revealing the limitations of existing supervision paradigms and offering insights into improving supervision signals.
    
    \item Extensive experiments demonstrate SpatialSV's strong cross-model and cross-dataset generalization in enhancing the spatial intelligence of MLLMs, and highlight its inherent interpretabiliy, as well as the potential for exploiting unlabeled visual data.
\end{itemize}

\begin{figure*}[!th]
\centering
    \includegraphics[width=0.985\textwidth]{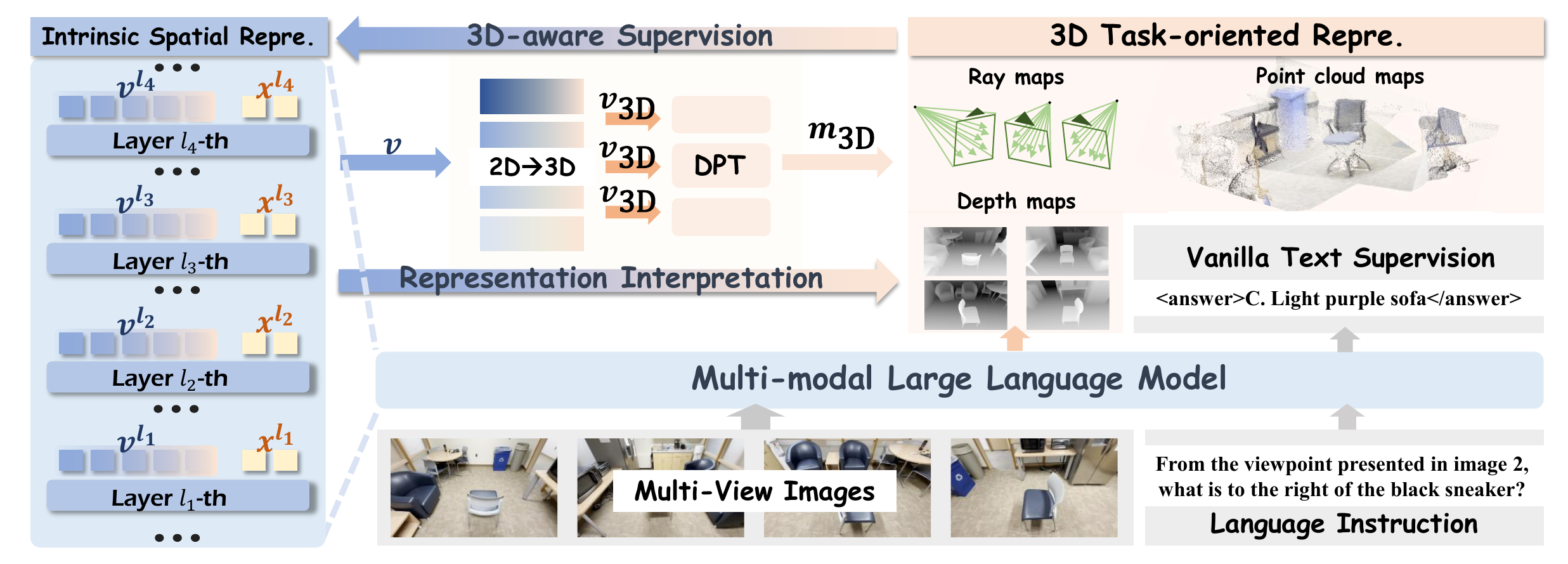}
    \caption{The schematic illustration of SpatialSV, a framework that internalizes interpretable and robust 3D spatial awareness in MLLMs via task-oriented visual supervision. We perform 2D-to-3D lifting of the MLLM's multi-layer hidden visual features and align them with explict 3D representations, including depth maps, ray maps, and point clouds. In this way, SpatialSV facilitates robust spatial representation learning and offers representation interpretability.}
    \label{SpatialSV}
\end{figure*}

\section{Method}

\subsection{Preliminaries}

A typical MLLM consists of a visual encoder $\mathcal{E}_{v}$ and a decoder-only large language model (LLM) $\mathcal{F}$. In the context of multi-view spatial understanding, the multimodal input comprises $N$ multi-view images $\mathcal{I}=\{I_{i}\}_{i=1}^{N}
$ along with a language instruction. The visual encoder transforms the images into visual features $\boldsymbol{v}^{0}=\mathcal{E}_{v}(\mathcal{I}) \in \mathbb{R}^{N_{v}\times d}$ aligned with the LLM's input space, where $N_v$ is the number of visual tokens, and $d$ is the feature dimension. The LLM then performs cross-modal feature interaction between the visual features and instruction tokens, producing $L$ layers of multimodal representations $\{\boldsymbol{v}^{i}, \boldsymbol{x}^{i}\}_{i=1}^{L}$, where $\boldsymbol{v}^{i}$ and $\boldsymbol{x}^{i}$ denote the $i$-th layer of visual and textual features, respectively. 

During the fine-tuning stage, the LLM applies causal modeling to the final-layer multimodal features and autoregressively predicts the next text token. The model is optimized by minimizing the standard cross-entropy loss:
\begin{equation}
    \mathcal{L}_{\text{text}} = - \sum_{t=1}^{T} \log p_{\theta}(\boldsymbol{x}^{L}_{t} \mid \boldsymbol{x}^{L}_{<t}, \boldsymbol{v}^{L}),
\end{equation}
where $p_{\theta}$ is the probability distribution conditioned on the previous multimodal context, and $T$ is the length of the text token sequence. Notably, only the textual features are explicitly supervised under this training paradigm. 

Following prior work \cite{huang2025mllms}, we inject spatial awareness into the MLLM by explicitly supervising its visual features. 
Specifically, we introduce a 3D vision foundation model $\mathcal{F}_{3D}$ to extract target features that encodes geometric priors, and apply 2D average pooling to match the size of MLLM's visual features: $\boldsymbol{v}_{\text{3D}}=
f_{\text{pool}}(\mathcal{F}_{\text{3D}}(\mathcal{I})) 
\in \mathbb{R}^{N_{v}\times d_{\text{3D}}}$,
where $d_{\text{3D}}$ is the dimension of target features.
In addition, we project MLLM's last-layer visual features into the target feature space to ensure dimension compatibility: 
$\boldsymbol{v}_{\text{proj}}=
\phi(\boldsymbol{v}^{L})
\in \mathbb{R}^{N_{v}\times d_{\text{3D}}}$.  
Finally, the model is optimized by maximizing the cosine similarity between $\boldsymbol{v}_{\text{proj}}$ and $\boldsymbol{v}_{\text{3D}}$ in a distillation paradigm:
\begin{equation}
    \mathcal{L}_{\text{distill}} = -S(\boldsymbol{v}_{\text{proj}}, \boldsymbol{v}_{\text{3D}}),
\end{equation}
where $S(\cdot,\cdot)$ denotes cosine similarity. However, this distillation supervision suffers from inherent limitations in the interpretability of representations. In contrast, an intuitive analysis of MLLMs' internal representation quality and its correlation with spatial intelligence is crucial for understanding the spatial mental modeling mechanism within MLLMs and uncovering their capability boundaries.

\subsection{Probe-based Spatial Representation Analysis}
\label{subsec:probe}
To acquire deeper insights into the intrinsic spatial intelligence of MLLMs, we investigate the correlation between representation quality and intelligence level under different supervision paradigms. Inspired by prior work \cite{el2024probing,chen2025feat2gs} which investigate the 3D-awareness within visual models, we introduce a probe-based representation analysis framework for MLLMs. 

Specifically, we evaluate multiple MLLMs, including Qwen2.5-VL \cite{bai2025qwen2}, LLaVA-NeXT-Video \cite{zhang2024video}, InternVL3 \cite{zhu2025internvl3}, and LLaVA-OneVision-1.5 \cite{li2024llava}, on the multi-view spatial understanding benchmark MindCube-Tiny \cite{yin2025spatial}. For each MLLM, we consider three variants: (i) an untuned version, (ii) a version fine-tuned solely with autoregressive text supervision, and (iii) a version that additionally incorporates visual distillation supervision beyond text supervision.

To quantify spatial intelligence, we compute each model's question-answering accuracy on MindCube-Tiny. To access the quality of intrinsic representations, we attach a DPT \cite{ranftl2021vision} module on top of MLLM's visual features for depth estimation. During this process, all MLLM parameters are frozen and only the DPT module is trainable. The depth map annotations for multi-view images in MindCube are obtained using DepthAnything-v3 \cite{lin2025depth}, and the DPT head is trained on the training split. After training, we evaluate the depth estimation quality on MindCube-Tiny using the RMSE metric, which serves as a proxy for the quality of MLLM's internal spatial representations. 

Figure \ref{Depth_Prob} presents both quantitative and qualitative results, from which we draw two key observations: (1) for a given MLLM, introducing visual distillation supervision substantially improves both intrinsic representation quality and spatial intelligence; (2) as the quality of intrinsic representations improves, the model’s spatial intelligence consistently increases. These findings strongly validate the effectiveness of 3D-aware visual supervision in enhancing the spatial intelligence of MLLMs. However, from the depth visualizations in Figure \ref{Depth_Prob}, we can observe a noticeable performance gap between the probing depth maps obtained under the distillation paradigm and the ground-truths. Specifically, the probing results appear overly blurred and suffer from significant loss of geometric details. This can be attributed to the inherent limitations of feature distillation supervision: (1) distillation at the feature level is inherently coarse-grained and lacks explicit, structured geometric constraints; (2) feature dimension alignment (e.g., via 2D average pooling) inevitably leads to information loss in the target representations. Collectively, these insights motivate a central question: \textbf{can we identify a fine-grained and intuitive form of 3D-aware visual supervision that can not only enhance but also interpret the intrinsic spatial intelligence of MLLMs?}

\begin{table*}[t]
\centering
\small
\renewcommand{\arraystretch}{1.0}

\begin{adjustbox}{width=\textwidth}
\begin{tabular}{lcccc|ccccccc}
\toprule
\multirow{2}{*}{\textbf{Method}} 
& \multicolumn{4}{c|}{\textbf{MindCube-Tiny}} 
& \multicolumn{7}{c}{\textbf{VSI-Bench}} \\
\cmidrule(lr){2-5} \cmidrule(lr){6-12}

& \textbf{Rotation} & \textbf{Among} & \textbf{Around} & \textcolor{brown}{\textbf{Overall}$\uparrow$}
& \makecell{\textbf{Rel.Dir.}\\\textbf{Hard}}
& \makecell{\textbf{Rel.Dir.}\\\textbf{Medium}}
& \makecell{\textbf{Rel.Dir.}\\\textbf{Easy}}
& \makecell{\textbf{Rel.}\\\textbf{Dist.}}
& \makecell{\textbf{App.}\\\textbf{Order}}
& \makecell{\textbf{Route}\\\textbf{Plan}}
& \textcolor{brown}{\textbf{Overall}$\uparrow$} \\

\midrule
\multicolumn{12}{c}{\textit{Baseline}} \\

Chance Level $(Random)$
& 26.5 & 24.5 & 23.8 & \textcolor{brown}{24.6}
& 26.5 & 25.7 & 24.9 & 24.1 & 23.3 & 28.4 & \textcolor{brown}{27.7} \\

Chance Level $(Frequency)$
& 34.5 & 34.8 & 33.3 & \textcolor{brown}{34.3}
& 25.2 & 33.6 & 50.2 & 25.1 & 25.2 & 29.4 & \textcolor{brown}{29.0} \\

\midrule
\multicolumn{12}{c}{\textit{Qwen2.5-VL Family}} \\

Qwen2.5-VL-3B 
& 34.5 & 36.0 & 32.3 & \textcolor{brown}{34.5}
& 34.3 & 33.3 & 29.0 & 31.7 & 22.7 & 27.8 & \textcolor{brown}{29.6} \\

Qwen2.5-VL-3B \textit{+ Text.}
& 33.5 & 49.7 & 74.8 & \textcolor{brown}{55.3}
& 35.7 & 35.7 & 26.3 & 31.1 & \textbf{23.6} & 29.9 & \textcolor{brown}{30.1} \\

Qwen2.5-VL-3B \textit{+ Distillation.}
& 34.5 & 51.5 & 81.3 & \textcolor{brown}{58.6}
& \textbf{35.9} & 34.1 & 30.9 & 32.1 & 22.8 & 32.0 & \textcolor{brown}{30.6} \\

\rowcolor{gray!15}
Qwen2.5-VL-3B \textit{+ SpatialSV}
& \textbf{36.0} & \textbf{56.2} & \textbf{84.5} & \textcolor{brown}{\textbf{62.3}}
& 34.6 & \textbf{37.8} & \textbf{34.6} & \textbf{33.5} & 23.3 & \textbf{37.1} & \textcolor{brown}{\textbf{32.2}} \\

Qwen2.5-VL-7B
& 35.0 & 29.7 & 26.8 & \textcolor{brown}{29.6}
& 28.4 & 24.9 & 45.2 & 31.8 & 29.5 & 31.4 & \textcolor{brown}{30.8} \\

Qwen2.5-VL-7B \textit{+ Text.}
& 35.5 & 51.0 & 73.5 & \textcolor{brown}{55.9}
& 27.4 & 31.2 & \textbf{51.2} & 31.4 & \textbf{30.1} & 34.5 & \textcolor{brown}{32.4} \\

Qwen2.5-VL-7B \textit{+ Distillation.}
& 35.0 & 53.8 & 76.5 & \textcolor{brown}{58.3}
& 29.0 & 32.5 & 50.2 & 34.8 & 29.1 & 37.6 & \textcolor{brown}{33.7} \\

\rowcolor{gray!15}
Qwen2.5-VL-7B \textit{+ SpatialSV}
& \textbf{37.5} & \textbf{58.5} & \textbf{81.8} & \textcolor{brown}{\textbf{62.8}}
& 29.5 & \textbf{35.2} & 48.4 & \textbf{37.2} & \textbf{30.1 }& \textbf{42.8} & \textcolor{brown}{\textbf{35.4}} \\

\midrule
\multicolumn{12}{c}{\textit{LLaVA-OneVision-1.5 Family}} \\

LLaVA-OneVision-1.5-8B
& 33.5 & 31.0 & 32.5 & \textcolor{brown}{32.0}
& 28.7 & 34.9 & 48.9 & 36.1 & 28.8 & 28.9 & \textcolor{brown}{33.5} \\

LLaVA-OneVision-1.5-8B \textit{+ Text.}
& 35.0 & 49.5 & 55.5 & \textcolor{brown}{49.1}
& 34.1 & 35.5 & \textbf{52.1} & 37.3 & \textbf{29.9} & 30.4 & \textcolor{brown}{35.5} \\

LLaVA-OneVision-1.5-8B \textit{+ Distillation.}
& 36.5 & 52.7 & 58.3 & \textcolor{brown}{51.8}
& 36.5 & 35.5 & 50.2 & 38.3 & 28.3 & 32.0 & \textcolor{brown}{35.7} \\

\rowcolor{gray!15}
LLaVA-OneVision-1.5-8B \textit{+ SpatialSV}
& \textbf{38.0} & \textbf{57.5} & \textbf{60.3} & \textcolor{brown}{\textbf{55.2}}
& \textbf{38.3} & \textbf{36.5} & 49.3 & \textbf{42.1} & 29.6 & \textbf{40.2} & \textcolor{brown}{\textbf{38.1}} \\

\midrule
\multicolumn{12}{c}{\textit{LLaVA-NeXT-Video Family}} \\

LLaVA-NeXT-Video-7B
& 34.5 & 41.7 & 34.5 & \textcolor{brown}{38.1}
& 27.1 & 32.5 & 50.7 & 35.9 & 27.5 & 31.4 & \textcolor{brown}{32.9} \\

LLaVA-NeXT-Video-7B \textit{+ Text.}
& 33.0 & 52.3 & 78.8 & \textcolor{brown}{57.9}
& 27.9 & \textbf{35.7} & 45.6 & 35.5 & 29.8 & 32.5 & \textcolor{brown}{33.6} \\

LLaVA-NeXT-Video-7B \textit{+ Distillation.}
& 36.5 & 55.2 & 79.5 & \textcolor{brown}{60.2}
& \textbf{29.5} & 33.6 & 48.4 & 37.5 & 29.5 & 34.5 & \textcolor{brown}{34.4} \\

\rowcolor{gray!15}
LLaVA-NeXT-Video-7B \textit{+ SpatialSV}
& \textbf{39.5} & \textbf{61.8} & \textbf{82.3} & \textcolor{brown}{\textbf{64.9}}
& 29.2 & 33.9 & \textbf{52.5} & \textbf{37.8} & \textbf{30.4} & \textbf{44.9} & \textcolor{brown}{\textbf{35.9}} \\

\midrule
\multicolumn{12}{c}{\textit{InternVL3 Family}} \\
InternVL3-2B 
& 31.0 & 33.3 & 24.8 & \textcolor{brown}{30.1}
& 26.5 & 31.5 & \textbf{48.4} & 30.1 & 24.4 & 34.0 & \textcolor{brown}{30.3} \\

InternVL3-2B \textit{+ Text.}
& 32.0 & 54.3 & 78.0 & \textcolor{brown}{58.5}
& 29.0 & 34.1 & 45.6 & 32.0 & \textbf{30.1} & 29.9 & \textcolor{brown}{32.4} \\

InternVL3-2B \textit{+ Distillation.}
& 35.0 & 56.5 & 79.5 & \textcolor{brown}{60.6}
& 29.5 & 33.3 & 47.0 & 33.0 & 28.0  & 31.4 & \textcolor{brown}{32.4} \\

\rowcolor{gray!15}
InternVL3-2B \textit{+ SpatialSV}
& \textbf{37.5} & \textbf{57.2} & \textbf{82.8} & \textcolor{brown}{\textbf{62.4}}
& \textbf{31.4} & \textbf{35.2} & 47.9 & \textbf{35.1} & 28.8 & \textbf{41.2} & \textcolor{brown}{\textbf{34.6}} \\

InternVL3-8B
& 34.5 & 39.3 & 33.3 & \textcolor{brown}{36.5}
& 23.6 & 34.1 & 46.1 & 34.1 & 32.9 & 32.5 & \textcolor{brown}{33.1} \\

InternVL3-8B \textit{+ Text.}
& 33.5 & 68.8 & 82.5 & \textcolor{brown}{67.5}
& 25.7 & 33.6 & 50.2 & 34.5 & 32.0 & 29.9 & \textcolor{brown}{33.5} \\

InternVL3-8B \textit{+ Distillation.}
& 35.5 & 70.8 & 81.5 & \textcolor{brown}{68.5}
& 28.4 & 36.0 & 47.5 & 35.5 & 31.7 & 34.0 & \textcolor{brown}{34.5} \\

\rowcolor{gray!15}
InternVL3-8B \textit{+ SpatialSV}
& \textbf{38.5} & \textbf{77.2} & \textbf{86.0} & \textcolor{brown}{\textbf{73.7}}
& \textbf{30.6} & \textbf{37.3} & \textbf{50.7} & \textbf{36.2} & \textbf{33.2} & \textbf{42.8} & \textcolor{brown}{\textbf{36.6}} \\

\bottomrule
\end{tabular}
\end{adjustbox}

\caption{Comparison of our approach (\textit{SpatialSV}) with the baseline models, the \textit{pure-text supervision} variants, and the \textit{feature distillation} varients on MindCube-Tiny and VSI-Bench. The best results within each model type are \textbf{bolded}. More detailed results across \textbf{8 MLLMs} are provided in \textit{Supp.C.1}.}
\label{tab:comparison_supervision}
\end{table*}

\subsection{SpatialSV}
To overcome the inherent limitations of feature distillation supervision, we propose SpatialSV, a framework that internalizes interpretable and robust 3D spatial awareness in MLLMs via task-oriented visual supervision, as depicted in Figure \ref{SpatialSV}.

3D VFMs \cite{wang2025vggt,lin2025depth} facilitate robust and geometrically consistent representation learning via overlapping yet complementary 3D downstream tasks, spanning point cloud reconstruction, camera pose estimation, depth estimation, and point tracking. Motivated by these work, we select three types of explicit spatial representations that are highly compatible with MLLM's visual feature maps, namely depth maps, ray maps and point cloud maps. These representations encode per-pixel spatial semantics, e.g., depth, camera pose and 3D coordinates, and are therefore more intuitive, structured and fine-grained than abstract feature maps. 
To obtain the ground-truth data, we leverage 3D VFMs to estimate multi-view depth maps and camera parameters, which are further transformed into ray maps and point clouds (see $\textit{Supp.A.1}$ for detailed computations). 
We align MLLM's visual features with these 3D task outputs to improve the quality of internal spatial representations.

Concretely, we employ a set of two-layer projectors $\Phi_{\text{3D}}$ to lift MLLM's multi-layer visual features into a shared 3D feature space. Among the hidden visual features of an MLLM, higher layers capture richer cross-modal semantics, whereas lower layers preserve more low-level visual information. Accordingly, we select visual features from four different layers to capture complementary spatial and semantic representations, with the set of selected layer indexes as 
$\{l_{i}\}_{i=1}^{4}$, and the projector set 
$\Phi_{\text{3D}}=\{\phi_{i}\}_{i=1}^{4}$. 
The projectors map the selected layer-wise features into multi-task shared 3D features:
\begin{equation}
    \boldsymbol{v}_{\text{3D}}^{i}=\phi_{i}(\boldsymbol{v}^{l_{i}}), i \in [1,4],
\end{equation}
where 
$\boldsymbol{v}_{\text{3D}}^{i} \in \mathbb{R}^{N_{v}\times d_{\text{3D}}^{i}}$ 
is the $i$-th selected layer of 3D-lifted features, and $d_{\text{3D}}^{i}$ is the corresponding dimension. 
Subsequently, we introduce task-decoupled DPT modules
$\mathcal{F}_{\text{DPT}}$
\cite{ranftl2021vision} which receive the lifted features as input and independently predict per-pixel depth, camera pose, and 3D coordinates:
\begin{equation}
    \{\boldsymbol{m}_{\text{3D}}^{t}, \boldsymbol{c}^{t}\}=
    \mathcal{F}_{\text{DPT}}^{t}(\{\boldsymbol{v}_{\text{3D}}^{i}\}_{i=1}^{4}),
\end{equation}
where $t\in \{\mathrm{depth},\mathrm{ray},\mathrm{pointcloud}\}$ denotes the task type; 
$\boldsymbol{m}_{\text{3D}}^{t}$ is the 3D outputs of task $t$; 
and $\boldsymbol{c}^{t}\in\mathbb{R}^{N_{v}}$ denotes the confidence of $\boldsymbol{m}_{\text{3D}}^{t}$. 
Specifically, 
$\boldsymbol{m}_{\text{3D}}^{\mathrm{dep.}} \in \mathbb{R}^{N_{v}\times 1}$ corresponds to per-pixel depth values; 
$\boldsymbol{m}_{\text{3D}}^{\mathrm{ray}} \in \mathbb{R}^{N_{v}\times 6}$ denotes per-pixel ray representations, where the first three channels encode the ray origin and the last three channels encode the ray direction; 
and  $\boldsymbol{m}_{\text{3D}}^{\mathrm{poi.}} \in \mathbb{R}^{N_{v}\times 3}$ represents per-pixel 3D spatial coordinates.
Following \cite{wang2025vggt,lin2025depth}, we then compute task-specific losses between the MLLM’s 3D-lifted predictions and the corresponding ground-truth annotations:
\begin{equation}
\begin{gathered}
    \mathcal{L}_{\text{3D}}^{t}=
    \mathcal{L}_{2}(\boldsymbol{y}_{\text{3D}}^{t},
    \boldsymbol{m}_{\text{3D}}^{t})+
    \mathcal{L}_{\text{conf}}(\boldsymbol{y}_{\text{3D}}^{t}, \boldsymbol{m}_{\text{3D}}^{t};\boldsymbol{c}^{t}))+ \\
    \mathds{1}_{\{t=\mathrm{dep.}\}}\,\mathcal{L}_{\text{grad}}+
    \mathds{1}_{\{t=\mathrm{poi.}\}}\,\mathcal{L}_{\text{norm}}, 
\end{gathered}
\end{equation}
where $\boldsymbol{y}_{\text{3D}}^{t}$ is the ground-truth of task $t$; $\mathds{I}_{\{\cdot\}}$ denotes the indicator function; $\mathcal{L}_{\text{conf}}$ is the confidence loss; $\mathcal{L}_{\text{grad}}$ is the gradient loss applied to the depth residual, penalizing blurred, misplaced, or spurious depth transitions; 
and $\mathcal{L}_{\text{norm}}$ is the surface normal loss that enforce the alignment between the local surface geometry of predicted points with the ground-truth. See $\textit{Supp.A.2}$ for the detailed loss computation procedure.

Consequently, the overall training objective integrates an  autoregressive text loss, a coarse-grained feature distillation loss, and fine-grained 3D-aware task-oriented losses:
\begin{equation}
    \mathcal{L}_{\text{total}}=
    \mathcal{L}_{\text{text}}+
    \mathcal{L}_{\text{distill}}+
    \mathcal{L}_{\text{3D}}^{\mathrm{dep.}}+
    \mathcal{L}_{\text{3D}}^{\mathrm{ray}}+
    \mathcal{L}_{\text{3D}}^{\mathrm{poi.}}.
\end{equation}
This approach combines coarse-grained guidance with fine-grained geometric constraints, which promotes robust intrinsic spatial representation learning within MLLMs. Moreover, during inference, the 3D-lifted predictions derived from the MLLM serve as an intuitive proxy to assess representation quality and delineate the model’s capability boundaries, as discussed in Sec. \ref{subsec:interpretability_ana}.

\begin{table*}[t]
\centering
\small
\renewcommand{\arraystretch}{1.0}

\resizebox{\textwidth}{!}{%
\begin{tabular}{lccccccccc}
\toprule
\multirow{2}{*}{\textbf{Method}} &
\makecell{\textbf{Ego3D-}\\\textbf{Bench}} & 
\textbf{Spatial457} & 
\makecell{\textbf{ViewSpatial-}\\\textbf{Bench}} & 
\makecell{\textbf{3DSR-}\\\textbf{Bench}} &
\textbf{SPBench} & 
\textbf{TopViewRS} & 
\textbf{CVBench} & 
\textbf{MMBench} & 

\textcolor{brown}{\textbf{Avg.}} \\
\midrule
Qwen2.5-VL-3B & 32.4 & 33.9 & 35.8 & 49.6 & 39.3 & 43.3 & 70.7 & 76.3 & \textcolor{brown}{47.7} \\
Qwen2.5-VL-3B+\textit{Text.} & 31.8 & 33.4 & 36.3 & 49.3 & 40.2 & 43.4 & 70.2 & 76.2 & \textcolor{brown}{47.6} \\

\rowcolor{gray!15}
Qwen2.5-VL-3B+\textit{SpatialSV} & \textbf{34.7} & \textbf{34.8} & \textbf{39.5} & \textbf{52.3} & \textbf{44.5} & \textbf{43.8} & \textbf{71.4} & \textbf{76.9} & \textcolor{brown}{\textbf{49.8}} \\

\midrule
Qwen2.5-VL-7B & 35.8 & 43.7 & 37.2 & 53.9 & 45.3 & 43.4 & 78.0 & 77.4 & \textcolor{brown}{51.8} \\
Qwen2.5-VL-7B+\textit{Text.} & 36.2 & 43.4 & 37.6 & 53.7 & 46.4 & 43.4 & 77.6 & 77.7 & \textcolor{brown}{52.0} \\

\rowcolor{gray!15}
Qwen2.5-VL-7B+\textit{SpatialSV} & \textbf{39.3} & \textbf{44.3} & \textbf{40.1} & \textbf{55.2} & \textbf{47.8} & \textbf{43.6} & \textbf{78.2} & \textbf{79.3} & \textcolor{brown}{\textbf{53.5}} \\

\bottomrule
\end{tabular}%
}
\caption{Comparison of our approach (\textit{SpatialSV}) with the baseline models and the \textit{pure-text supervision} variants on multiple spatial understanding and general benchmarks. The best results are \textbf{bolded}.}
\label{tab:comparison_benchmarks}
\end{table*}

\section{Experiments}

\subsection{Experimental Settings}
\noindent \textbf{Datasets}. For training, we use 10k training samples from MindCube \cite{yin2025spatial} to fine-tune the model. For evaluation, we focus on two benchmarks, MindCube-Tiny \cite{yin2025spatial} and VSI-Bench \cite{yang2025thinking}, both of which target spatial understanding from limited ego-centric views. 
We evaluate models on MindCube-Tiny and the multiple-choice split of VSI-Bench. 

\noindent \textbf{Evaluation Metrics}. We use Accuracy to evaluate the overall performance on multiple-choice questions, computed by exact matching between model predictions and ground-truths. For the 3D-lifted depth estimation task, we follow \cite{el2024probing} to report RMSE as the evaluation metric.

\noindent \textbf{Implementation Details}. We adopt multiple MLLMs as the baselines for SpatialSV, including Qwen2.5-VL \cite{bai2025qwen2}, 
LLaVA-OneVision-1.5 \cite{li2024llava}, 
LLaVA-NeXT-Video \cite{zhang2024video}, 
and InternVL3 \cite{zhu2025internvl3}. 
For all models, we fine-tune the vision–language projector, the large language model, all 2D-to-3D projectors, and the task-decoupled DPT modules, while freezing the visual encoder. 
All models are optimized using AdamW with a batch size of 16 and a warm-up ratio of 0.03. 
We use DepthAnything-v3 \cite{lin2025depth} by default to obtain 
the ground-truth supervision signals.
More detailed hyperparameter settings of all models are provided in $\textit{Supp.B.2}$.
All experiments are conducted on 8 NVIDIA H800 GPUs.

\subsection{Comparison with Baselines}
\noindent \textbf{Comparison with Other Supervision Paradigms}.
We evaluate our approach against the baselines, the pure-text supervision variants, and the feature distillation variants on 6 MLLMs spanning 4 model families (more detailed results present in \textit{Supp.C.1}). As shown in Table \ref{tab:comparison_supervision}, SpatialSV consistently achieves the best performance, demonstrating its strong \textbf{cross-model generalization} capability. Specifically, on MindCube-Tiny, SpatialSV yield overall gains ranging from 
$3.42\%$ to $12.66\%$ over the pure-text variants, and from $2.97\%$ to $7.81\%$ over the distillation variants. On VSI-Bench, we observe an overall improvement by $4.36\%$ to $6.79\%$ compared with the distillation variants. Notably, SpatialSV achieves an average improvement of $10.0\%$ over the pure-text variants on the \textit{relative distance} task, while on the more challenging \textit{route planning} task, the improvement reaches $32.6\%$. This indicates that SpatialSV effectively enhances the intrinsic spatial representations for 3D scenes within MLLMs, which is critical for estimating spatial attributes such as distance and direction.
Moreover, it is worth noting that no VSI-Bench-related sample is incorporated in training data; nevertheless, SpatialSV still achieves significant improvements, highlighting its remarkable \textbf{cross-dataset generalization} ability. 

\noindent \textbf{Comparison on Multiple Benchmarks}. We further evaluate our approach on several additional datasets, including 6 spatial understanding benchmarks, namely Ego3D-Bench \cite{gholami2025spatial}, Spatial457 \cite{wang2025spatial457}, ViewSpatial-Bench \cite{li2025viewspatial}, 3DSR-Bench \cite{ma20253dsrbench}, SP-Bench \cite{li2025spatialladder}, TopViewRS \cite{li2024topviewrs}, and 2 general benchmarks, CVBench \cite{zhu2025cvbench} and MMBench \cite{liu2024mmbench}. Detailed information about these datasets is provided in $\textit{Supp.B.1}$. As shown in Table 2, our method consistently outperforms baseline models and pure-text variants across all benchmarks, indicating that SpatialSV \textbf{enhances spatial intelligence while also preserving the general understanding capability}. This further underscores SpatialSV’s strong cross-dataset generalization.

\begin{figure}[t]
\centering
    \includegraphics[width=0.48\textwidth]{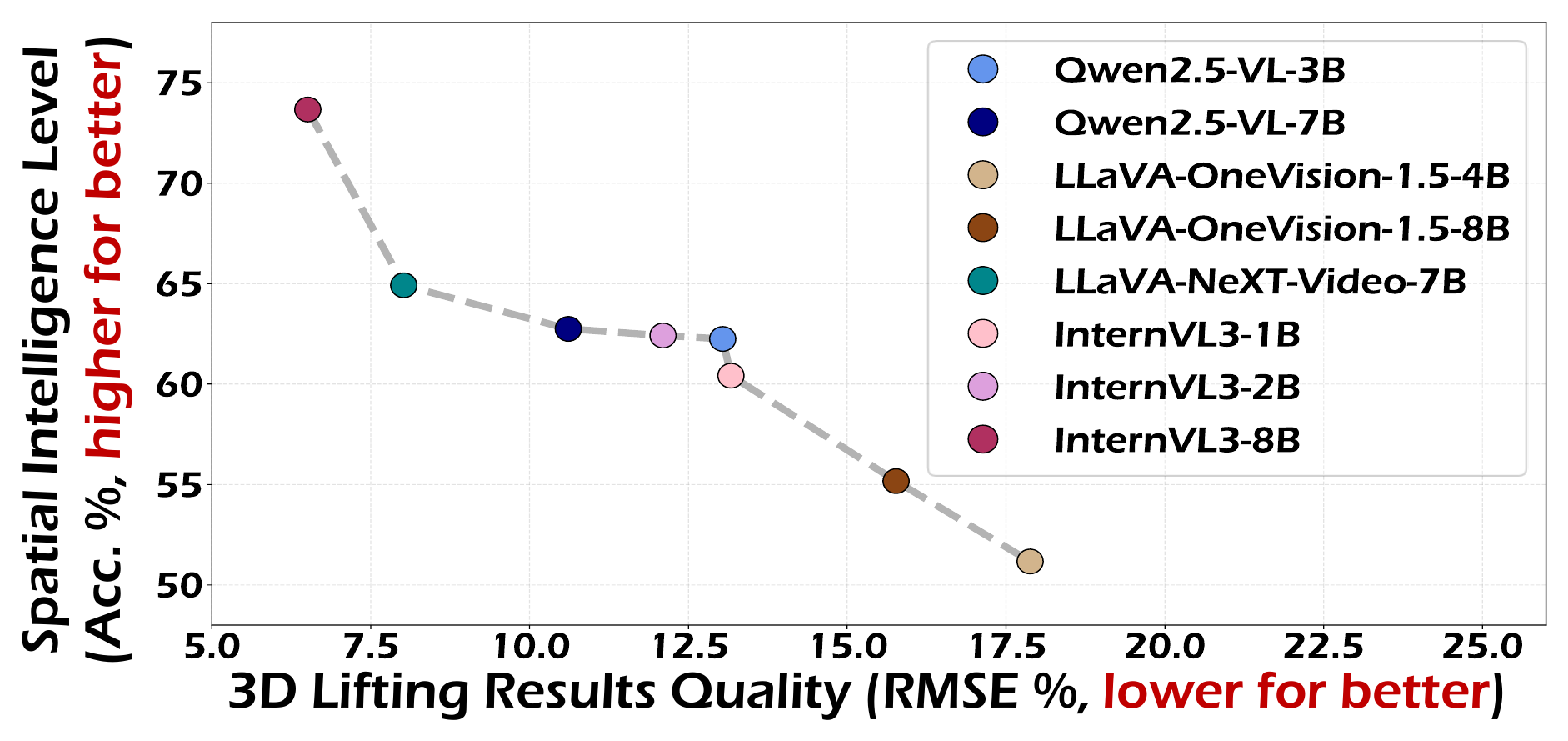}
    \caption{Correlation between the quality of the SpatialSV-based 3D lifting results and intra-model spatial intelligence level.}
    \label{Interpret_model}
\end{figure}

\begin{figure}[t]
\centering
    \includegraphics[width=0.47\textwidth]{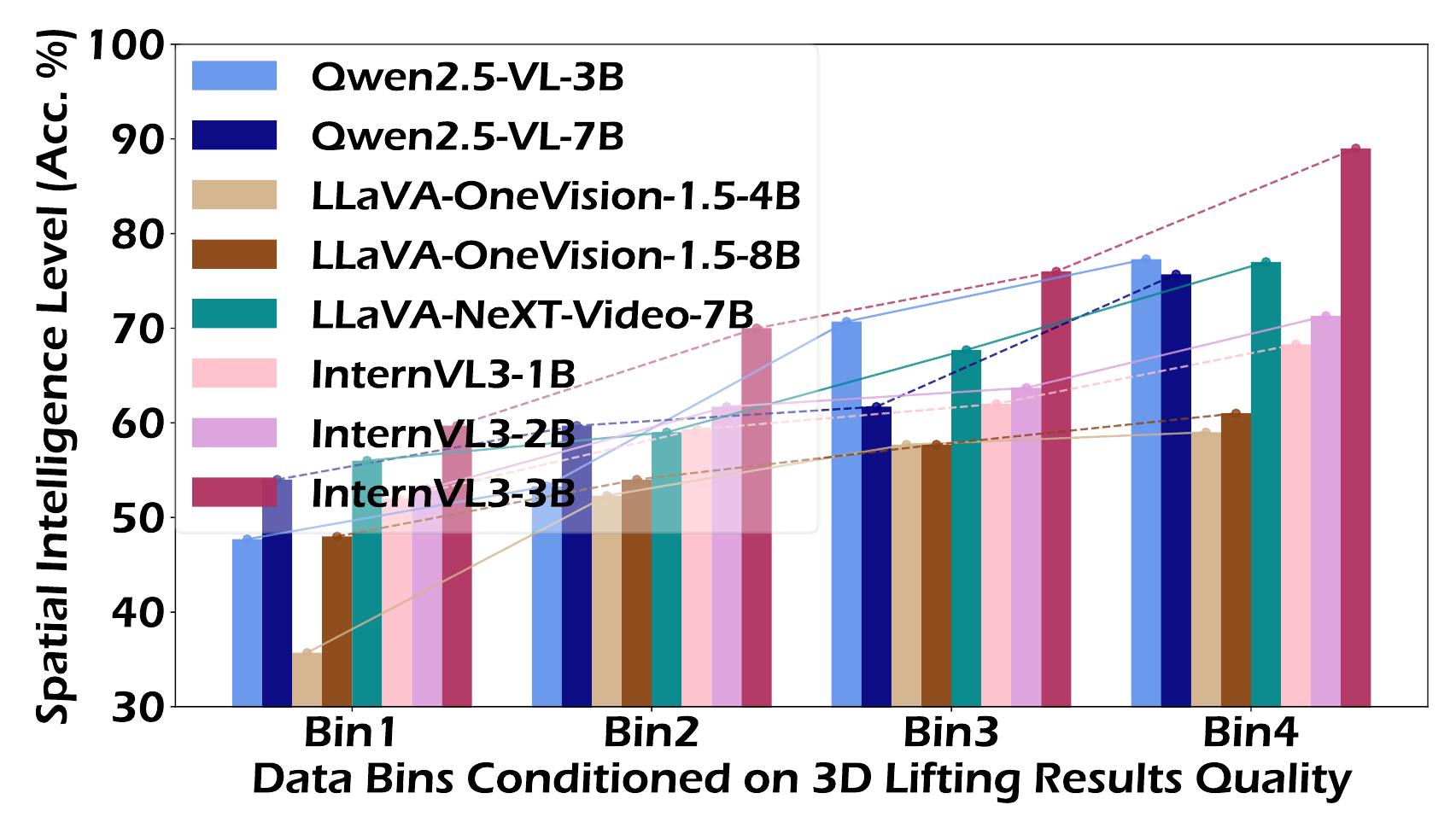}
    \caption{Correlation between the SpatialSV-based 3D lifting results with model-specific sample perferences. Samples are partitioned into 4 bins based on the quality of 3D lifting results.}
    \label{Interpret_data}
\end{figure}
\subsection{Interpretability Analysis}
\label{subsec:interpretability_ana}
In this section, we investigate the inherent interpretability of SpatialSV-based 3D lifting results.
We conduct both quantitative and qualitative analyses to reveal their correlations with intra-model spatial intelligence and model-specific sample preferences.

\begin{figure}[t]
\centering
    \includegraphics[width=0.48\textwidth]{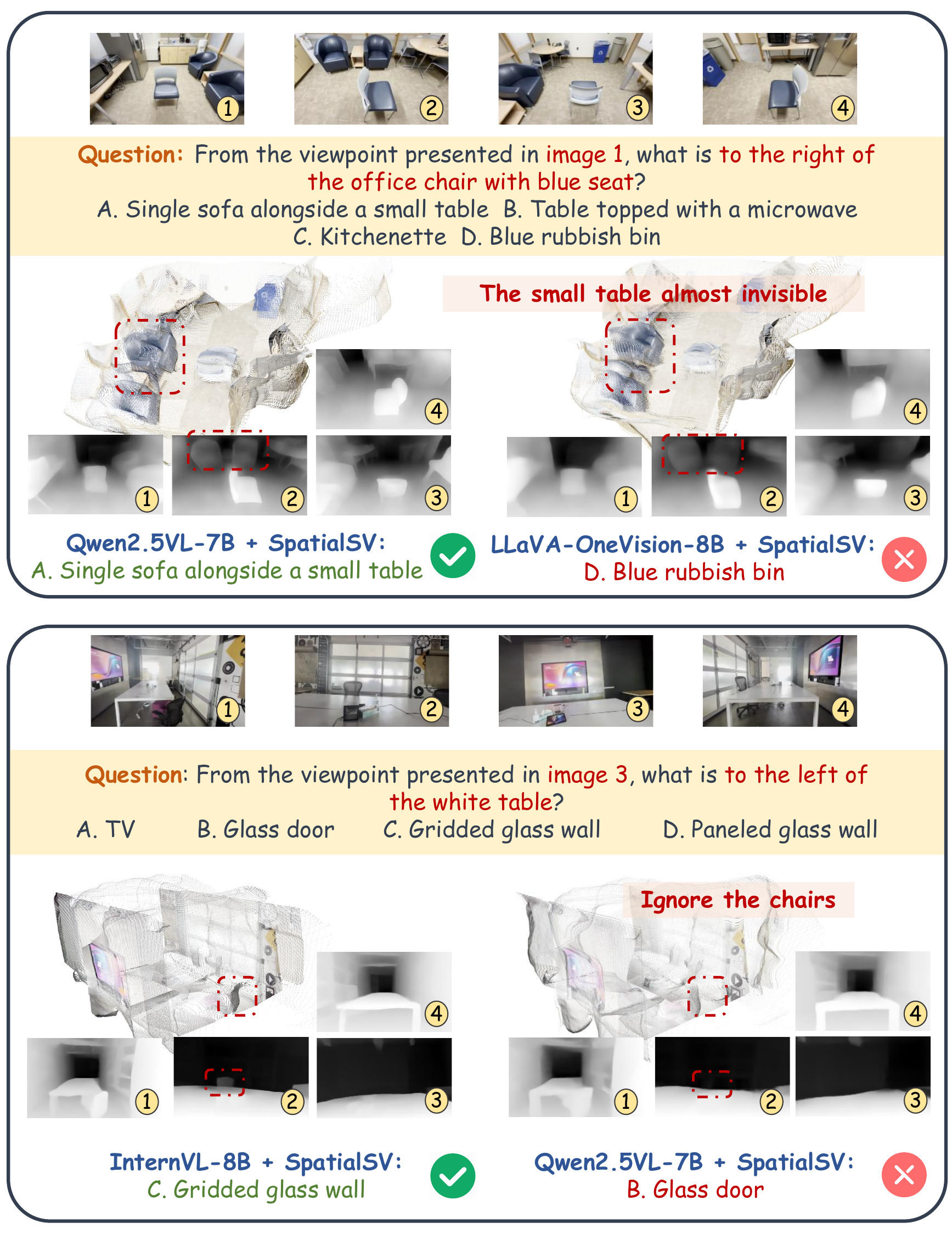}
    \caption{Qualitative results on MindCube-Tiny.}
    \label{Qualitative_results}
\end{figure}

\noindent \textbf{Correlation with Intra-Model Spatial Intelligence}. We use the depth estimation performance to quantify the quality of 3D lifting results, and VQA performance to measure MLLMs' intrinsic spatial intelligence, evaluating 8 MLLMs on MindCube-Tiny. In Figure \ref{Interpret_model}, we observe a strong correlation between the quality of 3D lifting results and intra-model spatial intelligence, which is aligned with the findings in Sec.\ref{subsec:probe}. Qualitatively, Figure \ref{Qualitative_results} shows that MLLMs exhibit varying representation capability under the same spatial context. In the first case, the depth maps and point clouds derived from LLaVA-OneVision-8B lack necessary details of the key object \textit{small table}, leading to its deviation from the correct option involving this object. In the second case, the failure of Qwen2.5-VL-7B can be partly attributed to its ignorance of \textit{chairs}, which could help establish cross-view correspondences.
In contrast, models that answer correctly can represent these critical objects in their 3D lifting results. These strongly validate the effectiveness of SpatialSV-based 3D lifting results as a proxy of intra-model spatial representations.

\noindent \textbf{Correlation with Model-Specific Sample Preferences}. For each MLLM, we partition the samples in  MindCube-Tiny into four bins according to the quality of estimated depth maps. From Bin1 to Bin4, the samples correspond to progressively high-quality 3D lifting results. In Figure \ref{Interpret_data}, QA accuracy exhibit a clear upward trend across all models, suggesting that \textbf{samples with unfaithful 3D lifting results tend to be more challenging for the models}. As shown in Figure 6, Qwen2.5-VL-7B succeeds in the first sample—accurately representing the spatial scene and providing the correct answer—but fails in the second sample in both aspects. This further highlights the correlation between SpatialSV-based 3D lifting results and model-specific sample preferences.

\subsection{Ablation Study}
\begin{table}[t]
    \centering
    \small
    \renewcommand{\arraystretch}{1.0}

    \resizebox{\columnwidth}{!}{%
    \begin{tabular}{cccccccc}
        \toprule
        \textbf{feat.} &
        \textbf{dep.} &
        \textbf{poi.} &
        \textbf{ray.} &
        \multicolumn{2}{c}{\textbf{Qwen2.5VL-3B}} &
        \multicolumn{2}{c}{\textbf{LLaVA-NV-7B}} \\
        
        \cmidrule(lr){5-6} \cmidrule(lr){7-8}
        
        & & & &
        \textbf{MC.} & \textbf{VSI.} &
        \textbf{MC.} & \textbf{VSI.} \\
        \midrule
        
        --&--&--&--& 55.3 & 30.1 & 57.9 & 33.6 \\
        \textbf{\cmark} &--&--&--& 57.4 & 30.2 & 60.3 & 34.0 \\
        --& \textbf{\cmark} &--&--& 59.4 & 31.3 & 62.2 & 34.8 \\
        --&--& \textbf{\cmark} &--& 58.2 & 31.5 & 62.4 & 33.9 \\
        --&--&--& \textbf{\cmark} & 56.3 & 30.0 & 60.2 & 34.2 \\
        --&
        \textbf{\cmark} & \textbf{\cmark} & \textbf{\cmark} & \underline{61.4} & \textbf{32.3} & \underline{63.7} & \underline{35.6} \\
        
        \textbf{\cmark} & \textbf{\cmark} & \textbf{\cmark} & \textbf{\cmark} & \textbf{62.3} & \underline{32.2} & \textbf{64.9} & \textbf{35.9} \\
        
        \bottomrule
    \end{tabular}%
    }
    \caption{Ablation of different supervision signals on MindCube-Tiny (MC.) and VSI-Bench (VSI.). \textit{Feat.}, \textit{dep}., \textit{poi.}, and  \textit{ray.} denote features, depth maps, ray maps and point cloud maps, respectively.}
    \label{tab:ablation_supervision_signal}
\end{table}

\begin{table}[t]
    \centering
    \small
    \renewcommand{\arraystretch}{1.0}

    \resizebox{\columnwidth}{!}{%
    \begin{tabular}{cccccc}
        \toprule
        \multirow{2}{*}{\makecell{\textbf{Layer}\\\textbf{Indexes}}} &
        \multirow{2}{*}{\makecell{\textbf{Projector}\\\textbf{Numbers}}} &
        \multicolumn{4}{c}{\textbf{MindCube-Tiny}} \\
        \cmidrule(lr){3-6}
        
        & &
        \textbf{Rotation} & \textbf{Among} &
        \textbf{Around} & \textbf{Overall} \\
        \midrule
        
        $(36,36,36,36)$ & 4 & 32.5 & 54.7 & 81.3 & 59.8 \\
        $(25,26,27,28)$ & 4 & \textbf{36.5} & 55.7 & 84.3 & 62.0 \\
        $(4,12,20,28)$ & 1 & 35.0 & 55.2 & 81.8 & 60.7 \\
        $(4,12,20,28)$ & 4 & 36.0 & \textbf{56.2} & \textbf{84.5} & \textbf{62.3} \\
        
        \bottomrule
    \end{tabular}%
    }
    \caption{Ablation of different visual layers and projector numbers on MindCube-Tiny for Qwen2.5-VL-3B. The best is \textbf{bolded}.}
    \label{tab:ablation_visual_layers}
\end{table}

\begin{table}[t]
    \centering
    \small
    \renewcommand{\arraystretch}{1.2}

    \resizebox{\columnwidth}{!}{%
    \begin{tabular}{cccccc}
        \toprule
        \multirow{2}{*}{\textbf{$50\%$ data}} &
        \multirow{2}{*}{\textbf{$50\%$ data}} &
        \multicolumn{4}{c}{\textbf{MindCube-Tiny}} \\
        \cmidrule(lr){3-6}
        
        & &
        \textbf{Rotation} & \textbf{Among} &
        \textbf{Around} & \textbf{Overall} \\
        \midrule
        
        $\mathcal{L}_{\text{text}}$ & -- & 27.5 & 42.0 & 64.8 & 47.2 \\
        $\mathcal{L}_{\text{text}}+\mathcal{L}_{\text{spatial}}$ & -- & 29.5 & 46.7 & 70.5 & 51.8 \\
        
        \rowcolor{gray!15}$\mathcal{L}_{\text{text}}+\mathcal{L}_{\text{spatial}}$ & $\mathcal{L}_{\text{spatial}}$ & 29.0 & 49.5 & 73.0 & 53.9 \\
        $\mathcal{L}_{\text{text}}$ & $\mathcal{L}_{\text{text}}$ & \textcolor{gray}{33.5} & \textcolor{gray}{49.7} & \textcolor{gray}{74.8} & \textcolor{gray}{55.3} \\
        
        \bottomrule
    \end{tabular}%
    }
    \caption{Semi-supervised learning with SpatialSV.}
    \label{tab:ablation_semi_supervised}
\end{table}

\noindent \textbf{Different 3D-Aware Visual Supervision Signals}. As reported in Table \ref{tab:ablation_supervision_signal}, we examine different 3D-aware visual supervision signals, including 3D VFM features, depth maps, point maps and ray maps. Each supervision signal yields performance gains to varying degrees, with depth maps contributing the most, likely due to their intuitive spatial semantics and relatively low alignment difficulty. 
When all task-oriented supervision signals are jointly applied, the model significantly outperforms any single-task supervision, highlighting \textbf{the effectiveness of overlapping yet complementary multi-task supervision for learning robust spatial representations}. Furthermore, incorporating VFM features brings additional gains, suggesting that coarse-grained feature alignment and fine-grained task-oriented spatial constraints can mutually guide and reinforce each other.  

\noindent \textbf{Different Supervised Visual Feature Layers and Numbers of 2D-to-3D Projectors}. As shown in Table \ref{tab:ablation_visual_layers}, we explore different configurations of visual feature layers and projector numbers on Qwen2.5-VL-3B, which contains 37 visual feature layers (e.g., $L=37$). Regarding the selection of visual layers, choosing $\{l_{i}\}_{i=1}^{4}=\{4,12,20,28\}$ yields the best performance, achieving a $4.18\%$ improvement over supervising only the final visual layer. 
This is because overly deep layers tend to lack low-level visual details crucial for 3D reconstruction, thereby hindering the learning of 3D spatial representations. 
In contrast, combining shallow visual details with deep cross-modal semantics which are complementary proves feasible. 
For projector numbers, employing layer-wise decoupled projectors leads to a $2.64\%$ performance gain compared to using a single shared projector. 
This can be attributed to the semantic heterogeneity across different visual layers, which necessitates distinct 2D-to-3D mapping functions.

\noindent \textbf{Semi-Supervised Learning with SpatialSV}. Given the scarcity of high-quality 3D vision-language data, we explore directly exploiting unlabeled visual data. Since task-oriented visual supervision is independent of text autoregressive supervision, SpatialSV is naturally suited to semi-supervised learning. Specifically, we split the training data into two non-overlapping subsets and consider 4 configurations: (i) applying pure-text supervision ($\mathcal{L}_{\text{text}}$) to $50\%$ data; 
(ii) applying $\mathcal{L}_{\text{text}}$ and SpatialSV 
($\mathcal{L}_{\text{spatial}}$)
to $50\%$ data; 
(iii) applying both losses to $50\%$ data and only $\mathcal{L}_{\text{spatial}}$ to the remaining unlabeled $50\%$;
(iv) applying $\mathcal{L}_{\text{text}}$ to $100\%$ data. 
From Table \ref{tab:ablation_semi_supervised}, We observe that the semi-supervised setting in (iii) achieves performance close to full text supervision ($53.9$ vs. $55.3$), while yielding a remarkable $14.19\%$ improvement over setting (i). These results \textbf{highly validate the potential of SpatialSV for effectively exploiting unlabeled raw visual data}. Notably, SpatialSV can directly leverage off-the-shelf 3D VFMs to generate 3D task annotations as semi-supervised signals, which is much less costly and easier to obtain than constructing spatial question-answering datasets.  

\section{Conclusion}
We propose SpatialSV, a novel framework that internalizes robust and interpretable 3D spatial awareness into MLLMs through task-oriented visual supervision. SpatialSV performs 2D-to-3D lifting of MLLMs' multi-layer visual features to align with explicit 3D representations, with the 3D-lifting results serving as intuitive and interpretable proxy of models' internal representations. 
Extensive experiments across multiple models and benchmarks demonstrate SpatialSV's effectiveness in enhancing and interpreting spatial intelligence. Semi-supervised learning results further validate SpatialSV's potential to leverage unlabeled visual data for scalable spatial representation learning.


\appendix
\section{Technical Details about SpatialSV}
\label{sec:a}

\subsection{Construction of Visual Supervision Signals}
\label{subsec:a1}
We leverage the off-the-shelf 3D vision foundation models (VFMs) to obtain 3D representations, including 3D feature maps, depth maps, ray maps and point cloud maps, to serve as visual supervision signals for SpatialSV. The 3D VFMs we use include VGGT \cite{wang2025vggt} and DepthAnything-v3 \cite{lin2025depth}. 
Specifically, we input multi-view images $\{I_{i}\}_{i=1}^{N}, I_{i} \in \mathbb{R}^{H\times W\times 3}$ into a 3D VFM $\mathcal{F}_{\text{3D}}$ to obtain 3D features, depth maps, and the camera parameters:
\begin{equation}
    (\{\boldsymbol{f}_{i}^{l}\}_{l=1}^{L_{\text{3D}}}, \boldsymbol{y}_{i}^{\mathrm{dep.}}, \boldsymbol{g}_{i})_{i=1}^{N_{v}}=
    \mathcal{F}_{\text{3D}}(\{I_{i}\}_{i=1}^{N_{v}}),
\end{equation}
where $\{\boldsymbol{f}_{i}^{l}\}_{l=1}^{L_{\text{3D}}}$ and $\boldsymbol{y}_{i}^{\mathrm{dep.}}\in \mathbb{R}^{H\times W\times 1}$ denote the 3D features and depth map corresponding to the $i$-th input image; $\boldsymbol{g}_{i}=[\textbf{q}_{i},\textbf{t}_{i},\textbf{f}_{i}]\in \mathbb{R}^{9}$ denotes the camera parameters, which is the concatenation of the rotation quaternion $\textbf{q}_{i}\in \mathbb{R}^{4}$, the translation vector $\textbf{t}_{i}\in \mathbb{R}^{3}$, and the field of view $\textbf{f}_{i}\in \mathbb{R}^{2}$. 
We can recover the the standard intrinsic matrix $\textbf{K}_{i}$ from $\textbf{f}_{i}$ and the principal point which is fixed at the image center $(\frac{W}{2},\frac{H}{2})$ by default, while the extrinsic parameters include $\textbf{R}_{i}=Rot(\textbf{q}_{i})$ and the camera center $\textbf{t}_i$.

Following \cite{huang2025mllms}, we extract the final-layer 3D features $\boldsymbol{f}^{L_{\text{3D}}}$ as distillation supervision signals. 
Due to the orthogonality constraint of rotation matrices which  may lead to unstable optimization, we do not directly lift the 2D visual features of MLLMs into camera parameters. Instead, we follow \cite{lin2025depth} to adopt ray maps as a substitute, which reduces the learning difficulty of the prediction task. 

To obtain the per-view ground-truth ray map $\boldsymbol{y}_{i}^{ray.}\in \mathbb{R}^{H\times W\times 6}
$
, we employ the camera parameters to compute the per-pixel ray direction:
\begin{equation}
    \textbf{d}_{i}=\textbf{R}_{i} \textbf{K}_{i}^{-1}\textbf{p}(u,v) \in\mathbb{R}^{H\times W\times3},
\end{equation}
where $\textbf{p}(u,v)=[u,v,1]^{T}$ is the homogeneous pixel coordinate, and the ray map $\boldsymbol{y}_{i}^{ray.}
=[\textbf{t}_{i}, \textbf{d}_{i}]$. Further, we can obtain the per-view ground-truth point cloud map via:
\begin{equation}
    \boldsymbol{y}_{i}^{\mathrm{poi.}}(u,v)=
    \textbf{t}_{i}+\boldsymbol{y}_{i}^{\mathrm{dep.}}(u,v)\cdot \textbf{d}_{i}(u,v).
\end{equation}
Consequently, the 3D outputs $\{\boldsymbol{y}_{i}^{\mathrm{dep.}}, \boldsymbol{y}_{i}^{\mathrm{ray.}}, \boldsymbol{y}_{i}^{\mathrm{poi.}}\}_{i=1}^{N}$ serve as the task-oriented visual supervision signals of SpatialSV.

\subsection{3D Task-Oriented Loss Functions}
\label{subsec:a2}
In this section, we detail the computation procedure of loss terms present in E.q. (5) of the main paper, namely the confidence loss $\mathcal{L}_{\text{conf}}$, the depth-map gradiant loss $\mathcal{L}_{\text{grad}}$, and the point-cloud normal loss $\mathcal{L}_{\text{norm}}$.

The confidence loss term aims to encourage high influence only at reliable regions and down-weight uncertain regions, which can be computed as:
\begin{equation}
\begin{gathered}
    \mathcal{L}_{\text{conf}}(\boldsymbol{y}_{\text{3D}}^{t}, \boldsymbol{m}_{\text{3D}}^{t};\boldsymbol{c}^{t}))=\\
    \frac{1}{|\Omega|}\sum_{i\in\Omega} 
    (||\boldsymbol{y}_{\text{3D}}^{t,i} - \boldsymbol{m}_{\text{3D}}^{t,i}||_2 \cdot \boldsymbol{c}^{t}
    - \lambda_{c} \cdot \log(\boldsymbol{c}^{t})),
\label{eq:LconfA}
\end{gathered}
\end{equation}
where $\Omega$ is the set of valid pixels; $||\cdot||_2$ is the $\mathcal{L}_{2}$ norm; and $\lambda_{c}$ is a regularization weight that prevent the model from predicting infinite uncertainty to minimize the loss.

The depth-map gradient loss encourages the predicted depth map to respect local depth variation, preserving edge structures and reducing depth ambiguity across neighbors, which can be formulated as:
\begin{equation}
    \mathcal{L}_{\text {grad }}=\sum_{k \in\{x, y\}} \frac{1}{|\Omega|} \sum_{i \in \Omega}\left(\left|\nabla_{k} \Delta_{i}\right| \cdot \boldsymbol{c}^{t}-\lambda_{c} \cdot \log \left(\boldsymbol{c}^{t}\right)\right)
\end{equation}
where $\Delta_{i}=\boldsymbol{y}_{\text{3D}}^{t,i} - \boldsymbol{m}_{\text{3D}}^{t,i}$ denotes the difference map; $\nabla_{x}$ and $\nabla_{y}$ are horizontal and vertical spatial gradients, which ensure the local structure of the prediction matches the ground-truth even if the absolute depth is incorrect.

The point-cloud normal loss enforces geometry consistency of the reconstructed surface by aligning per-pixel surface normals derived from the prediction with the ground-truth, which can be computed via:
\begin{equation}
\begin{gathered}
    \mathcal{L}_{\text {normal }}= \frac{1}{|\Omega|} \sum_{i \in \Omega}(
    (1-\left\langle\mathbf{n}(\boldsymbol{m}_{\text{3D}}^{t})_{i}, \mathbf{n}(\boldsymbol{y}_{\text{3D}}^{t})_{i}\right\rangle) \cdot\boldsymbol{c}^{t}
    - \\ \lambda_{c} \cdot \log (\boldsymbol{c}^{t}))
\end{gathered}
\end{equation}
where $\langle\cdot,\cdot\rangle$ denotes the dot product, with the term $1-\langle\textbf{n}_{1},\textbf{n}_{2}\rangle$ approximating $1-cos(\theta)$ and penalizing angular deviation.

\section{Implementation Details of Experiments}
\label{sec:b}

\subsection{Dataset Details}
\label{subsec:b1}
In this section, we detail the information of all datasets used in our experiments, including eight spatial understanding benchmarks, namely MindCube \cite{yin2025spatial}, VSI-Bench \cite{yang2025thinking}, Ego3D-Bench \cite{gholami2025spatial}, Spatial457 \cite{wang2025spatial457}, ViewSpatial-Bench \cite{li2025viewspatial}, 3DSR-Bench \cite{ma20253dsrbench}, SP-Bench \cite{li2025spatialladder}, TopViewRS \cite{li2024topviewrs}, along with two general benchmarks, CVBench \cite{zhu2025cvbench} and MMBench \cite{liu2024mmbench}.

\textbf{MindCube} targets key challenges such as cross-view object consistency and reasoning about occluded or invisible objects, and comprises three types of camera motion: \textit{Rotation}, \textit{Among}, and \textit{Around}. In our experiments, we use the 10k training samples in MindCube to fine-tune the models, and leverage the 1.2k samples in MindCube-Tiny for evaluation.

\textbf{VSI-Bench} emphasizes the understanding of spatial relations and multi-object correspondence, covering spatial relation tasks (e.g., relative distance and relative direction), spatiotemporal tasks (e.g., object appearance order), and complex spatial manipulation tasks (e.g., route planning). We use its multiple-choice set for evaluation.

\textbf{Ego3D-Bench} is designed to evaluate spatial reasoning from ego-centric, multi-view outdoor data. It comprises over 8.6k QA pairs across five tasks spanning absolute distance, relative distance, localization, motion reasoning, and travel time. We use its multiple-choice set for evaluation.

\textbf{Spatial457} is a synthetic benchmark that emphasizes four core capabilities spanning multi-object recognition, 2D locations, 3D locations, and 3D orientation. It introduces three spatial relation types—2D spatial relations in the camera view, 6D spatial relations that combine 3D position and orientation from a target object’s perspective, and collision relations for forward/backward motion—captured in seven question types across five difficulty levels.

\textbf{ViewSpatial-Bench} focuses on multi-perspective spatial localization. It comprises 5.7k QA pairs across 1.3k scenes, covering task types that test spatial localization from both camera and human viewpoints. 

\textbf{3DSR-Bench} features 12 question types that probe 3D properties such as object height, inter-object location, orientation, and multi-object relations, emphasizing 3D grounding, camera extrinsics, object poses, depth, and multi-object reasoning.

\textbf{SP-Bench} is a two-part dataset consisting of SPBench-SI (Single-Image) and SPBench-MV (Multi-View). It targets four spatial reasoning tasks—absolute distance, relative distance, object size, and relative direction—with questions provided in numerical and multiple-choice formats. We employ SPBench-SI for evaluation.

\textbf{TopViewRS} targets top-view spatial understanding through 11k multiple-choice questions posed on indoor top-view maps. It defines four tasks spanning top-view recognition, top-view localization, static spatial seasoning and dynamic spatial reasoning.

\textbf{CVBench} is designed to evaluate cross-video relational reasoning, comprising 1.3k diversified videos accompanied by 1k carefully crafted multiple-choice QA pairs that span three hierarchical tiers: cross-video object association, cross-video event association and cross-video complex reasoning.

\textbf{MMBench} comprises 3.2k data samples across 20 leaf abilities organized under a hierarchical taxonomy that spans perception (coarse and fine-grained, including single- and cross-instance) and reasoning (e.g., attribute, spatial, social, nature, physical, logic, plus complex structuralized image-text understanding and future prediction).

\subsection{Detailed Hyperparameter Settings}
\label{subsec:b2}
We apply SpatialSV to 8 MLLMs across 4 model families, including Qwen2.5-VL-3B, Qwen2.5-VL-7B \cite{bai2025qwen2}, LLaVA-NeXT-Video-7B \cite{zhang2024video}, LLaVA-OneVision-1.5-4B, LLaVA-OneVision-1.5-8B \cite{li2024llava}, InternVL3-1B, InternVL3-2B, and InternVL3-8B \cite{zhu2025internvl3}. For all models, we fine-tune the vision–language projector, the large language model, all 2D-to-3D projectors, and the task-decoupled DPT modules, while freezing the visual encoder. 
All models are optimized using AdamW with a batch size of 16 and a warm-up ratio of 0.03. Table \ref{tab:hyperparameters} presents the settings of learning rates and supervised visual layer indexes for each model. All experiments are conducted on 8 NVIDIA H800 GPUs.

\begin{table}[t]
    \centering
    \small
    \setlength{\tabcolsep}{2pt}
    \renewcommand{\arraystretch}{1.2}
    \begin{tabular}{lcccc}
        \toprule
        $\textbf{Model}$ & $\boldsymbol{lr}_{\text{llm}}$ & $\boldsymbol{lr}_{\text{mm}}$ & $\boldsymbol{lr}_{\text{3D}}$ & $\textbf{Supervised Visual Layers}$ \\
        \midrule
        Qwen2.5-VL-3B  & 2e-7 & 1e-5 & 1e-5 & $(4,12,20,28)$ \\
        Qwen2.5-VL-7B  & 2e-7 & 1e-5 & 1e-5 & $(3,10,17,24)$ \\
        LLaVA-OV-1.5-4B  & 2e-7 & 1e-5 & 1e-5 & $(4,12,20,28)$ \\
        LLaVA-OV-1.5-8B  & 2e-7 & 1e-5 & 1e-5 & $(4,12,20,28)$ \\
        LLaVA-NV-7B & 1e-5 & 1e-5 & 1e-5 & $(4,11,18,25)$ \\
        InternVL3-1B & 1e-5 & 1e-5 & 1e-5 & $(3,9,15,21)$ \\
        InternVL3-2B & 1e-5 & 1e-5 & 1e-5 & $(3,9,15,21)$ \\
        InternVL3-3B & 1e-5 & 1e-5 & 1e-5 & $(3,9,15,21)$ \\
        
        \bottomrule
    \end{tabular}
    \caption{The settings of learning rates and supervised visual layers for different MLLMs. $\boldsymbol{lr}_{\text{llm}}$, $\boldsymbol{lr}_{\text{mm}}$, and $\boldsymbol{lr}_{\text{3D}}$ denote the learning rates of the language model, the vision-language projector, and the 2D-to-3D components including projectors and DPT heads, respectively.}
    \label{tab:hyperparameters}
\end{table}

\section{More Experimental Results}
\label{sec:c}

\subsection{Comparison with Other Supervision Methods}
\label{subsec:c1}
As shown in Table \ref{tab:comparison_supervision_method} , we present a detailed comparison with other supervision paradigms, especially the distillation paradigm based on 3D VFM features \cite{huang2025mllms}, across 8 MLLMs on MindCube-Tiny and VSI-Bench. 
Notably, SpatialSV consistently outperforms both the pure-text and feature distillation variants, indicating that SpatialSV can inject robust spatial awareness into the models via 3D task-oriented fine-grained spatial constraints.

\subsection{Ablation on the Selection of 3D VFMs}
\label{subsec:c2}
In this section, we conduct an ablation on the selection of 3D VFMs that are leveraged to obtain the visual supervision signals. As shown in Table \ref{tab:ablation_VFM}, we compare the performance of Qwen2.5-VL-3B and LLaVA-NeXT-Video-7B when VGGT and DepthAnything-v3 are respectively used as sources of the target representations. Both configurations exhibit comparable performance, indicating the robustness of our approach to the source of supervision signals.

\subsection{More Qualitative Results}
\label{subsec:c3}
\begin{figure}[t]
\captionsetup{skip=0pt}
\centering
    \includegraphics[width=0.48\textwidth]{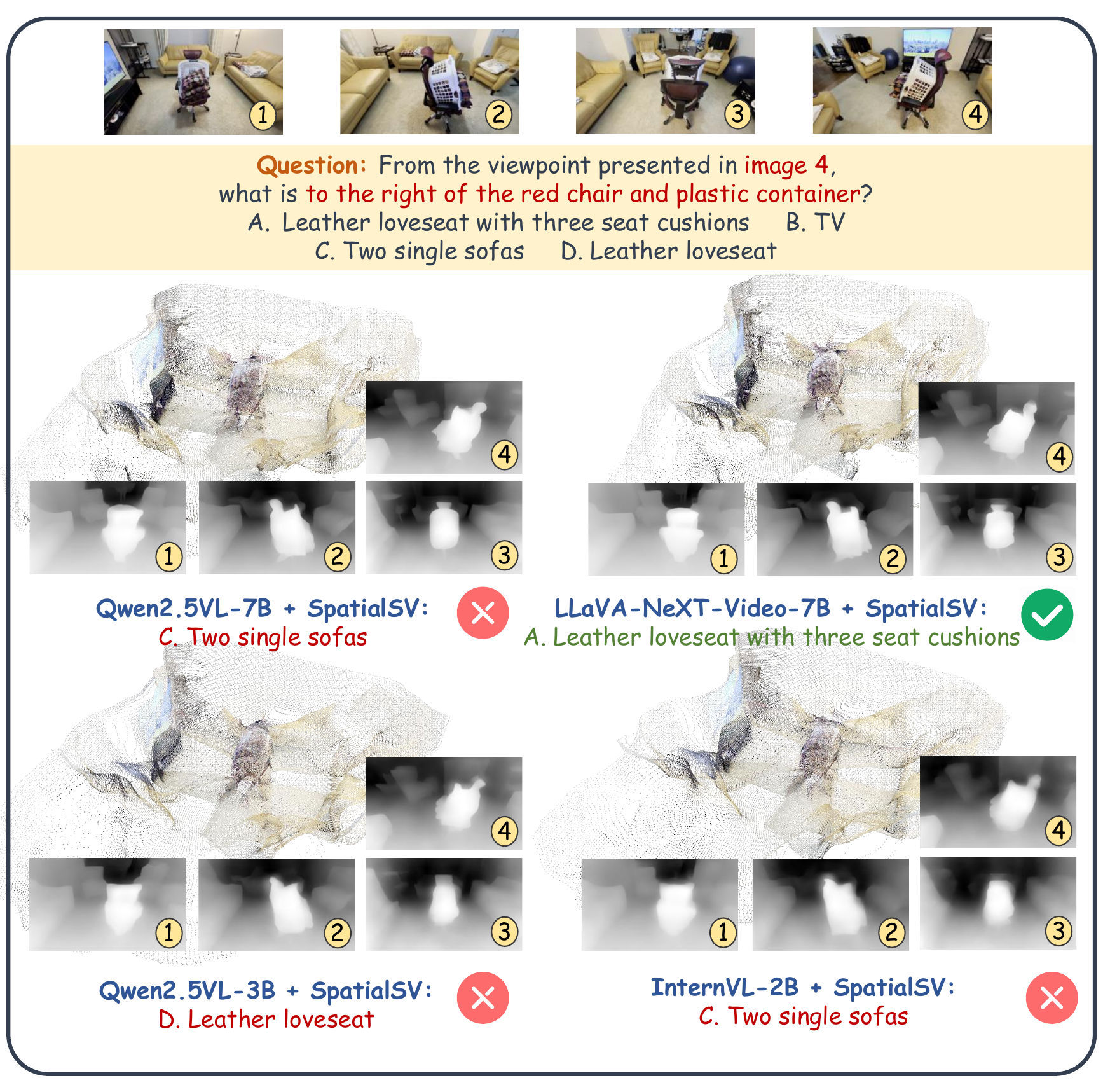}
    \caption{Qualitative results on MindCube-Tiny (Example1).}
    \label{vis_supp_1}
\end{figure}

\begin{figure}[t]
\captionsetup{skip=0pt}
\centering
    \includegraphics[width=0.48\textwidth]{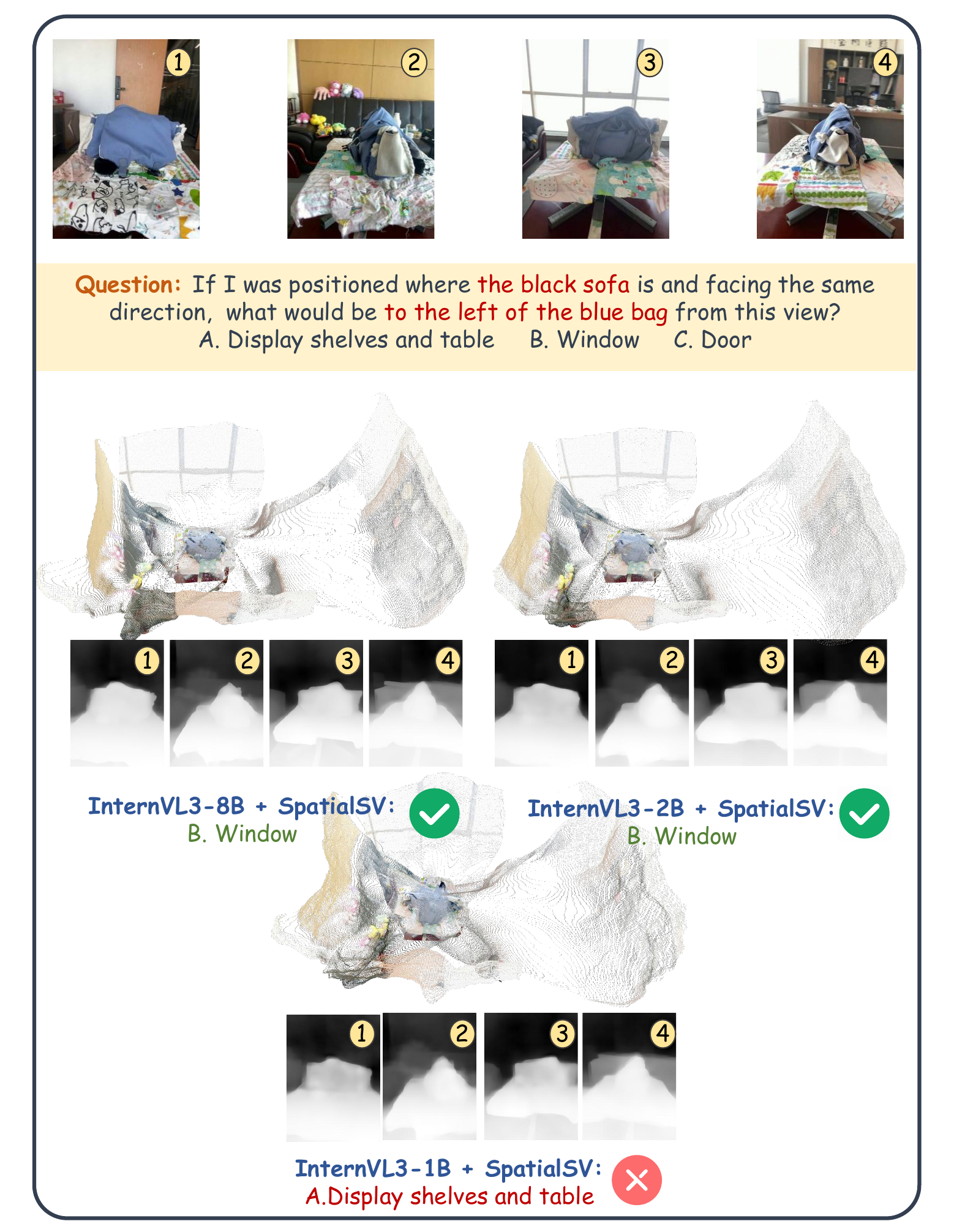}
    \caption{Qualitative results on MindCube-Tiny (Example2).}
    \label{vis_supp_2}
\end{figure}

\begin{figure}[t]
\captionsetup{skip=0pt}
\centering
    \includegraphics[width=0.48\textwidth]{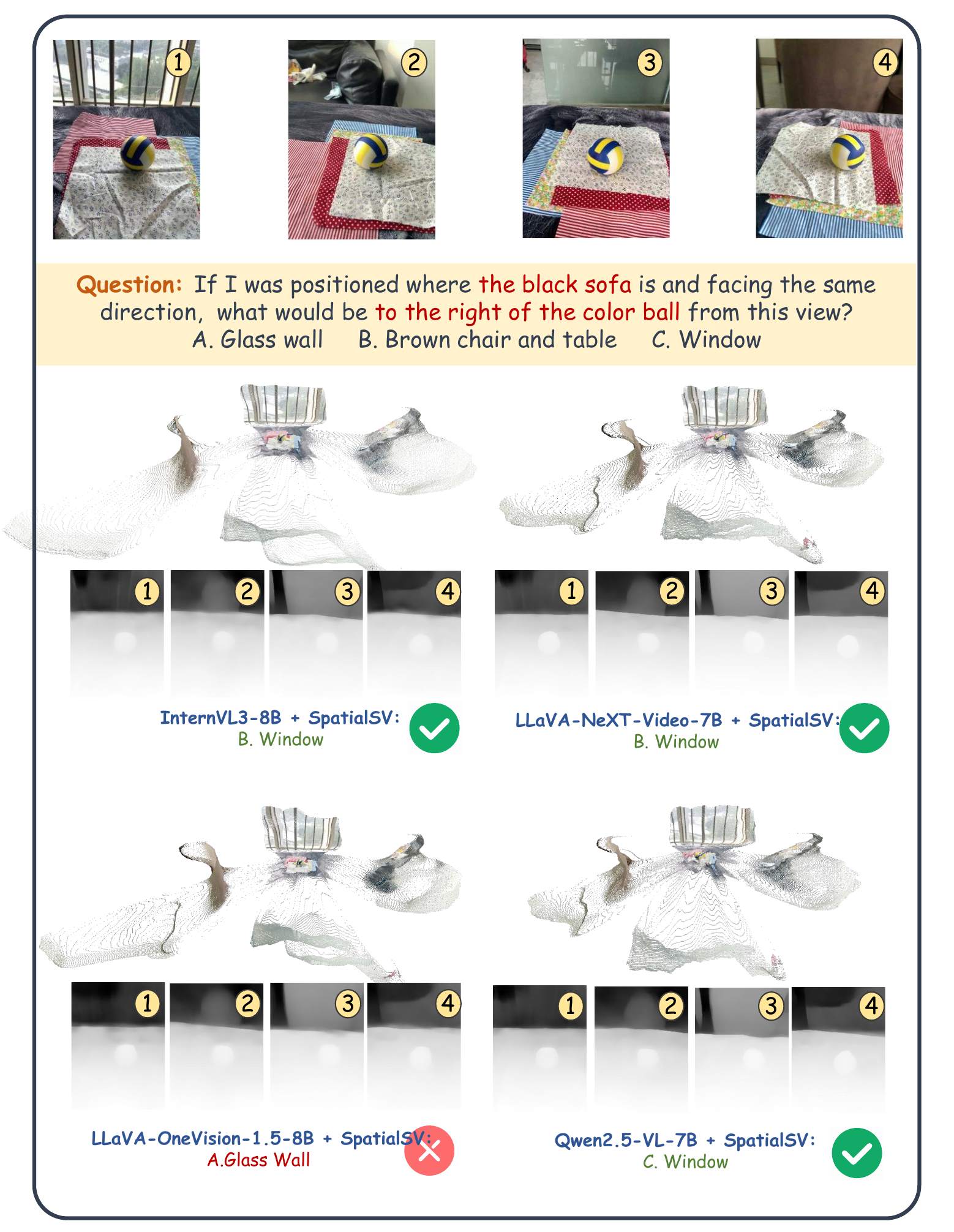}
    \caption{Qualitative results on MindCube-Tiny (Example3).}
    \label{vis_supp_3}
\end{figure}

We provide additional qualitative results that demonstrate varying internal representation capabilities and VQA performance of different MLLMs on specific samples. The qualitative results are shown in Figure \ref{vis_supp_1}-\ref{vis_supp_3}.

\begin{table}[t]
    \centering
    \small
    \setlength{\tabcolsep}{2pt}
    \renewcommand{\arraystretch}{1.0}
    \begin{tabular}{cccccc}
        \toprule
        \multirow{2}{*}{$\textbf{MLLM}$} &
        \multirow{2}{*}{$\textbf{3D VFM}$} &
        \multicolumn{4}{c}{\textbf{MindCube-Tiny}} \\
        \cmidrule(lr){3-6}
        
        & &
        \textbf{Rotation} & \textbf{Among} &
        \textbf{Around} & \textbf{Overall} \\
        \midrule

        \multirow{2}{*}{Qwen2.5VL-3B} & 
        VGGT & 34.5 & 56.8 & 83.5 & 62.0 \\

        & DAMv3 & 36.0 & 56.2 & 84.5 & 62.3 \\
        \midrule

        \multirow{2}{*}{LLaVA-NV-7B} & 
        VGGT & 38.0 & 62.8 & 82.8 & 65.3 \\
        & DAMv3 & 39.5 & 61.8 & 82.3 & 64.9 \\        
        \bottomrule
    \end{tabular}
    \captionsetup{font=small}
    \caption{Ablation on the selection of 3D VFMs.}
    \label{tab:ablation_VFM}
\end{table}

\begin{table*}[!th]
\centering
\small
\setlength{\tabcolsep}{4pt}
\renewcommand{\arraystretch}{1.2}

\begin{adjustbox}{width=\textwidth}
\begin{tabular}{lcccc|ccccccc}
\toprule
\multirow{2}{*}{\textbf{Method}} 
& \multicolumn{4}{c|}{\textbf{MindCube-Tiny}} 
& \multicolumn{7}{c}{\textbf{VSI-Bench}} \\
\cmidrule(lr){2-5} \cmidrule(lr){6-12}

& \textbf{Rotation} & \textbf{Among} & \textbf{Around} & \textcolor{brown}{\textbf{Overall}$\uparrow$}
& \makecell{\textbf{Rel.Dir.}\\\textbf{Hard}}
& \makecell{\textbf{Rel.Dir.}\\\textbf{Medium}}
& \makecell{\textbf{Rel.Dir.}\\\textbf{Easy}}
& \makecell{\textbf{Rel.}\\\textbf{Dist.}}
& \makecell{\textbf{App.}\\\textbf{Order}}
& \makecell{\textbf{Route}\\\textbf{Plan}}
& \textcolor{brown}{\textbf{Overall}$\uparrow$} \\

\midrule
\multicolumn{12}{c}{\textit{Baseline}} \\

Chance Level $(Random)$
& 26.5 & 24.5 & 23.8 & \textcolor{brown}{24.6}
& 26.5 & 25.7 & 24.9 & 24.1 & 23.3 & 28.4 & \textcolor{brown}{27.7} \\

Chance Level $(Frequency)$
& 34.5 & 34.8 & 33.3 & \textcolor{brown}{34.3}
& 25.2 & 33.6 & 50.2 & 25.1 & 25.2 & 29.4 & \textcolor{brown}{29.0} \\

\midrule
\multicolumn{12}{c}{\textit{Qwen2.5-VL Family}} \\

Qwen2.5-VL-3B 
& 34.5 & 36.0 & 32.3 & \textcolor{brown}{34.5}
& 34.3 & 33.3 & 29.0 & 31.7 & 22.7 & 27.8 & \textcolor{brown}{29.6} \\

Qwen2.5-VL-3B \textit{+ Text.}
& 33.5 & 49.7 & 74.8 & \textcolor{brown}{55.3}
& 35.7 & 35.7 & 26.3 & 31.1 & \textbf{23.6} & 29.9 & \textcolor{brown}{30.1} \\

Qwen2.5-VL-3B \textit{+ Distillation.}
& 34.5 & 51.5 & 81.3 & \textcolor{brown}{58.6}
& \textbf{35.9} & 34.1 & 30.9 & 32.1 & 22.8 & 32.0 & \textcolor{brown}{30.6} \\

\rowcolor{gray!15}
Qwen2.5-VL-3B \textit{+ SpatialSV}
& \textbf{36.0} & \textbf{56.2} & \textbf{84.5} & \textcolor{brown}{\textbf{62.3}}
& 34.6 & \textbf{37.8} & \textbf{34.6} & \textbf{33.5} & 23.3 & \textbf{37.1} & \textcolor{brown}{\textbf{32.2}} \\

Qwen2.5-VL-7B
& 35.0 & 29.7 & 26.8 & \textcolor{brown}{29.6}
& 28.4 & 24.9 & 45.2 & 31.8 & 29.5 & 31.4 & \textcolor{brown}{30.8} \\

Qwen2.5-VL-7B \textit{+ Text.}
& 35.5 & 51.0 & 73.5 & \textcolor{brown}{55.9}
& 27.4 & 31.2 & \textbf{51.2} & 31.4 & \textbf{30.1} & 34.5 & \textcolor{brown}{32.4} \\

Qwen2.5-VL-7B \textit{+ Distillation.}
& 35.0 & 53.8 & 76.5 & \textcolor{brown}{58.3}
& 29.0 & 32.5 & 50.2 & 34.8 & 29.1 & 37.6 & \textcolor{brown}{33.7} \\

\rowcolor{gray!15}
Qwen2.5-VL-7B \textit{+ SpatialSV}
& \textbf{37.5} & \textbf{58.5} & \textbf{81.8} & \textcolor{brown}{\textbf{62.8}}
& 29.5 & \textbf{35.2} & 48.4 & \textbf{37.2} & \textbf{30.1 }& \textbf{42.8} & \textcolor{brown}{\textbf{35.4}} \\

\midrule
\multicolumn{12}{c}{\textit{LLaVA-OneVision-1.5 Family}} \\

LLaVA-OneVision-1.5-4B
& 26.5 & 35.2 & 32.3 & \textcolor{brown}{32.8}
& 37.3 & \textbf{35.5} & 46.1 & 37.0 & 24.3 & 29.9 & \textcolor{brown}{33.9} \\

LLaVA-OneVision-1.5-4B \textit{+ Text.}
& \textbf{35.0} & 49.0 & 38.3 & \textcolor{brown}{43.1}
& 38.9 & 34.1 & 48.9 & 38.6 & 25.1 & 30.9 & \textcolor{brown}{34.9} \\

LLaVA-OneVision-1.5-4B \textit{+ Distillation.}
& 34.5  & 51.2 & 42.3 & \textcolor{brown}{45.4}
& 40.0 & \textbf{35.5} & 47.9 & 39.0 & 24.6 & 33.0 & \textcolor{brown}{35.3} \\

\rowcolor{gray!15}
LLaVA-OneVision-1.5-4B \textit{+ SpatialSV}
& 34.5 & \textbf{56.5} & \textbf{51.5} & \textcolor{brown}{\textbf{51.2}}
& \textbf{40.5} & 35.2 & \textbf{52.5} & \textbf{42.0} & \textbf{25.4} & \textbf{36.1} & \textcolor{brown}{\textbf{37.1}} \\

LLaVA-OneVision-1.5-8B
& 33.5 & 31.0 & 32.5 & \textcolor{brown}{32.0}
& 28.7 & 34.9 & 48.9 & 36.1 & 28.8 & 28.9 & \textcolor{brown}{33.5} \\

LLaVA-OneVision-1.5-8B \textit{+ Text.}
& 35.0 & 49.5 & 55.5 & \textcolor{brown}{49.1}
& 34.1 & 35.5 & \textbf{52.1} & 37.3 & \textbf{29.9} & 30.4 & \textcolor{brown}{35.5} \\

LLaVA-OneVision-1.5-8B \textit{+ Distillation.}
& 36.5 & 52.7 & 58.3 & \textcolor{brown}{51.8}
& 36.5 & 35.5 & 50.2 & 38.3 & 28.3 & 32.0 & \textcolor{brown}{35.7} \\

\rowcolor{gray!15}
LLaVA-OneVision-1.5-8B \textit{+ SpatialSV}
& \textbf{38.0} & \textbf{57.5} & \textbf{60.3} & \textcolor{brown}{\textbf{55.2}}
& \textbf{38.3} & \textbf{36.5} & 49.3 & \textbf{42.1} & 29.6 & \textbf{40.2} & \textcolor{brown}{\textbf{38.1}} \\

\midrule
\multicolumn{12}{c}{\textit{LLaVA-NeXT-Video Family}} \\

LLaVA-NeXT-Video-7B
& 34.5 & 41.7 & 34.5 & \textcolor{brown}{38.1}
& 27.1 & 32.5 & 50.7 & 35.9 & 27.5 & 31.4 & \textcolor{brown}{32.9} \\

LLaVA-NeXT-Video-7B \textit{+ Text.}
& 33.0 & 52.3 & 78.8 & \textcolor{brown}{57.9}
& 27.9 & \textbf{35.7} & 45.6 & 35.5 & 29.8 & 32.5 & \textcolor{brown}{33.6} \\

LLaVA-NeXT-Video-7B \textit{+ Distillation.}
& 36.5 & 55.2 & 79.5 & \textcolor{brown}{60.2}
& \textbf{29.5} & 33.6 & 48.4 & 37.5 & 29.5 & 34.5 & \textcolor{brown}{34.4} \\

\rowcolor{gray!15}
LLaVA-NeXT-Video-7B \textit{+ SpatialSV}
& \textbf{39.5} & \textbf{61.8} & \textbf{82.3} & \textcolor{brown}{\textbf{64.9}}
& 29.2 & 33.9 & \textbf{52.5} & \textbf{37.8} & \textbf{30.4} & \textbf{44.9} & \textcolor{brown}{\textbf{35.9}} \\

\midrule
\multicolumn{12}{c}{\textit{InternVL3 Family}} \\

InternVL3-1B
& 31.5 & 37.3 & 32.0 & \textcolor{brown}{34.6}
& \textbf{30.8} & 33.9 & 47.0 & 26.6 & 24.6 & 33.5 & \textcolor{brown}{30.2} \\

InternVL3-1B \textit{+ Text.}
& 33.5 & 52.3 & \textbf{80.0} & \textcolor{brown}{58.4}
& 29.0 & 34.7 & 44.7 & 27.3 & \textbf{25.4} & 30.9 & \textcolor{brown}{30.0} \\

InternVL3-1B \textit{+ Distillation.}
& \textbf{35.5} & 53.5 & 77.0  & \textcolor{brown}{58.3}
& 29.8 & \textbf{35.2} & 43.8 & \textbf{28.2} & 25.1 & 32.0 & \textcolor{brown}{30.4} \\

\rowcolor{gray!15}
InternVL3-1B \textit{+ SpatialSV}
& 35.0 & \textbf{56.8} & 78.5 & \textcolor{brown}{\textbf{60.4}}
& 30.0 & 34.7 & \textbf{50.2} & \textbf{28.2} & 24.9 & \textbf{38.7} & \textcolor{brown}{\textbf{31.4}} \\

InternVL3-2B 
& 31.0 & 33.3 & 24.8 & \textcolor{brown}{30.1}
& 26.5 & 31.5 & \textbf{48.4} & 30.1 & 24.4 & 34.0 & \textcolor{brown}{30.3} \\

InternVL3-2B \textit{+ Text.}
& 32.0 & 54.3 & 78.0 & \textcolor{brown}{58.5}
& 29.0 & 34.1 & 45.6 & 32.0 & \textbf{30.1} & 29.9 & \textcolor{brown}{32.4} \\

InternVL3-2B \textit{+ Distillation.}
& 35.0 & 56.5 & 79.5 & \textcolor{brown}{60.6}
& 29.5 & 33.3 & 47.0 & 33.0 & 28.0  & 31.4 & \textcolor{brown}{32.4} \\

\rowcolor{gray!15}
InternVL3-2B \textit{+ SpatialSV}
& \textbf{37.5} & \textbf{57.2} & \textbf{82.8} & \textcolor{brown}{\textbf{62.4}}
& \textbf{31.4} & \textbf{35.2} & 47.9 & \textbf{35.1} & 28.8 & \textbf{41.2} & \textcolor{brown}{\textbf{34.6}} \\

InternVL3-8B
& 34.5 & 39.3 & 33.3 & \textcolor{brown}{36.5}
& 23.6 & 34.1 & 46.1 & 34.1 & 32.9 & 32.5 & \textcolor{brown}{33.1} \\

InternVL3-8B \textit{+ Text.}
& 33.5 & 68.8 & 82.5 & \textcolor{brown}{67.5}
& 25.7 & 33.6 & 50.2 & 34.5 & 32.0 & 29.9 & \textcolor{brown}{33.5} \\

InternVL3-8B \textit{+ Distillation.}
& 35.5 & 70.8 & 81.5 & \textcolor{brown}{68.5}
& 28.4 & 36.0 & 47.5 & 35.5 & 31.7 & 34.0 & \textcolor{brown}{34.5} \\

\rowcolor{gray!15}
InternVL3-8B \textit{+ SpatialSV}
& \textbf{38.5} & \textbf{77.2} & \textbf{86.0} & \textcolor{brown}{\textbf{73.7}}
& \textbf{30.6} & \textbf{37.3} & \textbf{50.7} & \textbf{36.2} & \textbf{33.2} & \textbf{42.8} & \textcolor{brown}{\textbf{36.6}} \\

\bottomrule
\end{tabular}
\end{adjustbox}

\caption{Detailed comparison of our approach (\textit{SpatialSV}) with the baseline models, the \textit{pure-text supervision} variant, and the \textit{feature distillation} variant on MindCube-Tiny and VSI-Bench across 8 MLLMs. The best results within each model type are \textbf{bolded}.}
\label{tab:comparison_supervision_method}
\end{table*}

\section{Related Work}
\label{sec:related_work}
The significant advancements of multimodal large language models \cite{bai2025qwen2,zhu2025internvl3,zhou2026logic} in visual understanding and reasoning has motivated growing efforts to endow them with strong \textbf{Spatial Intelligence}.
which is crucial for real-world applications such as autonomous driving \cite{zhou2025towards,huang2025robotron,tang2026letp} and embodied interaction \cite{hu20253dllm,song2025robospatial,yang20253d}. 
According to the way of incorporating spatial knowledge, existing methods can be broadly categorized into two lines of research: external spatial prior–based approaches and internalized spatial awareness–based approaches.

\textbf{External spatial prior–based approaches} aim to compensate for the limited intrinsic spatial awareness of MLLMs by constructing external inputs that encode rich spatial priors. 
Among these, prompt-based methods rely on auxiliary models or tools to generate various forms of linguistic or visual prompts, including 
chains of thought \cite{zhang2025spatial}, 
cognitive maps \cite{yang2025thinking,gholami2025spatial}, 
object markers \cite{qi2025gpt4scene}, 
bird’s-eye-view (BEV) maps \cite{zhu2025struct2d}, 
motion trajectories \cite{li2025see}, and 
3D bounding boxes \cite{liu2025abstract}. 
For example, 
Ego3D-VLM \cite{gholami2025spatial} employs off-the-shelf referring expression detection and depth estimation models to construct cognitive maps as model inputs. 
MindJourney \cite{yang2025mindjourney} leverages multi-round interactions between a video diffusion model and a VLM to generate viewpoints containing spatial evidence as prompts. 
However, such methods rely heavily on the performance of external models, making them sensitive to upstream errors while significantly increasing inference overhead. 
Another line of approaches within this paradigm explicitly incorporate 3D inputs through multimodal fusions. 
\cite{xu2024pointllm,chen2024ll3da}
require explicit point cloud encoding and alignment with the textual modality, 
while 
\cite{zheng2025video,zhu2025llava}
transform point clouds into 3D positional encodings. 
\cite{cheng2024spatialrgpt,liu2025ssr}
encode depth maps as additional inputs. 
However, these methods rely on highly customized modules and precise cross-modal alignment while also increases computational costs, posing challenges for practical deployment.

\textbf{Internalized spatial awareness–based approaches} emphasize the learning of internal spatial representation within MLLMs. 
Early approaches \cite{chen2024spatialvlm} leverage large-scale 2D spatial vision–language datasets and train models solely with textual supervision, which proves insufficient for injecting spatial awareness due to the absence of 3D prior guidance. 
Ross3D \cite{wang2025ross3d} introduces masked cross-view and global-view reconstruction in the latent space as auxiliary supervision, which remains confined to the 2D domain and struggle with the learning of 3D representations. 
More recent approaches \cite{huang2025mllms,chen2025think}, employ representations from 3D vision foundation models (VFMs) \cite{wang2025vggt,lin2025depth} as supervision signals for feature distillation. 
Nevertheless, imitation at the feature level is inherently coarse-grained and lacks structured and explicit geometric constraints. 
Moreover, the intrinsic representations learned by these methods remain uninterpretable, 
which hinders a deeper understanding of the spatial mental modeling mechanisms within MLLMs. 

Our approach follows the second paradigm, focusing on internalizing spatial awareness in MLLMs to maintain a streamlined model architecture and facilitate practical deployment. More importantly, we introduce explicit 3D representations as supervision signals, enabling task-oriented, fine-grained spatial constraints while preserving the interpretability of the internalized spatial awareness.

\bibliographystyle{named}
\bibliography{ijcai26}

\end{document}